\documentclass[runningheads,dvipsnames,table,xcdraw]{llncs}

\usepackage{eccv}

\usepackage{eccvabbrv}

\usepackage{graphicx}
\usepackage{booktabs}
\usepackage{bm}

\usepackage[accsupp]{axessibility}  %

\usepackage[breaklinks,colorlinks,citecolor=black,linkcolor=black]{hyperref}

\usepackage{orcidlink}

\usepackage{enumitem}

\usepackage{algpseudocode}
\definecolor{commgreen}{HTML}{009933}
\usepackage[linesnumbered,ruled,vlined]{algorithm2e}

\SetCommentSty{licommfont}

\setenumerate[1]{itemsep=0pt,partopsep=0pt,parsep=\parskip,topsep=2.5pt}
\setitemize[1]{itemsep=1pt,partopsep=0pt,parsep=\parskip,topsep=1.5pt}
\setdescription{itemsep=0pt,partopsep=0pt,parsep=\parskip,topsep=2.5pt}
\usepackage{booktabs}
\usepackage{multirow}
\usepackage{dsfont}
\usepackage{threeparttable}
\usepackage{amsmath}
\usepackage{amssymb}
\usepackage{microtype}
\usepackage{mathtools}
\usepackage{pifont}
\usepackage[geometry]{ifsym}

\usepackage[capitalize]{cleveref}
\Crefname{section}{Section}{Sections}
\crefname{section}{Sec.}{Secs.}
\Crefname{table}{Table}{Tables}
\crefname{table}{Tab.}{Tabs.}

\newcommand{\mathvec}[1]{\bm{#1}}
\newcommand\bdtitle[1]{\noindent\textbf{#1}}
\newcommand\topr{\toprule[1.2pt]}
\newcommand\toprr{\toprule[0.85pt]}
\newcommand\midr{\toprule[0.65pt]}
\newcommand\bottomr{\bottomrule[1.2pt]}
\renewcommand{\arraystretch}{1.05}

\usepackage{adjustbox}
\newcommand{\graynum}[1]{\textcolor{figgray}{\ding{#1}}}
\newlength{\compactlen}
\definecolor{tabgreen}{RGB}{59,125,35}
\definecolor{figgray}{RGB}{127,127,127}
\definecolor{tabred}{RGB}{192,0,0}
\definecolor{cvprblue}{rgb}{0.21,0.49,0.74}
\definecolor{bondiblue}{rgb}{0.0, 0.18, 0.65}
\definecolor{dodgerblue}{rgb}{0.12, 0.56, 1.0}
\definecolor{jade}{HTML}{0EB83A}
\definecolor{bluearrow}{HTML}{306EBA}

\definecolor{fig2g}{RGB}{74, 177, 61}
\definecolor{fig2b}{RGB}{68, 125, 216}
\definecolor{fig2r}{RGB}{242, 118, 132}

\newcommand{\tabup}[1]{\textcolor{bondiblue!80}{\scriptsize (\textbf{#1})}}
\newcommand{\tabdown}[1]{\textcolor{bondiblue!80}{\scriptsize (\textbf{#1})}}

\newcommand{\tabequ}[1]{\textcolor{bondiblue!80}{\scriptsize (\textbf{#1})}}

\newcommand{\ssymbol}[1]{^{\@fnsymbol{#1}}}

\usepackage{rotating}

\begin{document}

\newcommand{\validset}{{\textit{validation} set}}
\newcommand{\trainset}{{\textit{training} set}}
\newcommand{\testset}{{\textit{test} set}}
\newcommand{\pddshort}{{R\kern-0.1em A\kern-0.1em PiD}}
\newcommand{\rpddfull}{{Intra-Ring {\pddfull}}}
\newcommand{\cpddfull}{{Intra-Class {\pddfull}}}
\newcommand{\rpdd}{{R-{\pddshort}}}
\newcommand{\cpdd}{{C-{\pddshort}}}
\newcommand{\ourmodel}{{\pddshort}-Seg}
\newcommand{\pddfull}{{Range-Aware Pointwise Distance Distribution}}

\title{\textls[-50]{{\ourmodel}: {\pddfull} Networks for 3D LiDAR Segmentation}} 

\titlerunning{{\pddshort}-Seg Networks for 3D LiDAR Segmentation}

\author{Li Li\inst{1}\orcidlink{0000-0002-9392-7862} \and
Hubert P. H. Shum\inst{1}\orcidlink{0000-0001-5651-6039} \and
Toby P. Breckon\inst{1,2}\orcidlink{0000-0003-1666-7590}}

\authorrunning{L.~Li et al.}

\institute{Department of \{Computer Science$^{1}$ $|$ Engineering$^{2}$\}, Durham University, UK \\
\email{\{li.li4,\ hubert.shum,\ toby.breckon\}@durham.ac.uk}}

\maketitle

\vspace{-15pt}
\begin{abstract}
    \noindent
\textls[-4]{
3D point clouds play a pivotal role in outdoor scene perception, especially in the context of autonomous driving. Recent advancements in 3D LiDAR segmentation often focus intensely on the spatial positioning and distribution of points for accurate segmentation. However, these methods, while robust in variable conditions, encounter challenges due to sole reliance on coordinates and point intensity, leading to poor isometric invariance and suboptimal segmentation.
To tackle this challenge, our work introduces \textbf{R}ange-\textbf{A}ware \textbf{P}o\textbf{i}ntwise \textbf{D}istance Distribution ({\pddshort}) features and the associated {\ourmodel} architecture. Our {\pddshort} features exhibit rigid transformation invariance and effectively adapt to variations in point density, with a design focus on capturing the localized geometry of neighboring structures. They utilize inherent LiDAR isotropic radiation and semantic categorization for enhanced local representation and computational efficiency, while incorporating a 4D distance metric that integrates geometric and surface material reflectivity for improved semantic segmentation. %
To effectively embed high-dimensional {\pddshort} features, we propose a double-nested autoencoder structure with a novel class-aware embedding objective to encode high-dimensional features into manageable voxel-wise embeddings. Additionally, we propose {\ourmodel} which incorporates a channel-wise attention fusion and two effective {\pddshort}-Seg variants, further optimizing the embedding for enhanced performance and generalization.
Our method outperforms contemporary LiDAR segmentation work in terms of mIoU on SemanticKITTI ($\mathbf{76.1}$) and nuScenes ($\mathbf{83.6}$) datasets. 
}


\vspace{-20pt}

\end{abstract}
\vspace{-5pt}

\section{Introduction}

LiDAR-based point clouds, characterized by pointwise 3D positions and LiDAR intensity/reflectivity~\cite{Geiger2012,li2021durlara,sun2020scalabilitya,nuscenes2019}, play a pivotal role in outdoor scene understanding, particularly in perception systems for autonomous driving.
Recent literature in the domain of 3D LiDAR segmentation has witnessed significant strides, with myriad approaches attempting pointwise 3D semantic scene labeling~\cite{sun2024empirical}. Leveraging color~\cite{af2s3net,liu2023segment,liu2023uniseg}, range imagery~\cite{kong2023rethinking,liong2020amvnet}, and Birds Eye View (BEV) projections~\cite{ye2023lidarmultinet,liong2020amvnet}, multi-modal methods integrate diverse data streams from LiDAR and other sensors to enhance feature representation and segmentation accuracy. Single-modal (LiDAR-only) methods~\cite{hou2022pointtovoxel, li2023less, rpvnet, zhu2021cylindrical} effectively exploit spatial data interpretation and avoid multi-modal issues, \eg, modality heterogeneity and limited sensor field of view intersection~\cite{chen2023svqnet}.

\begin{center}
    \centering
    \vspace{-20pt}
    \captionsetup{type=figure}
    \includegraphics[width=1.005\linewidth]{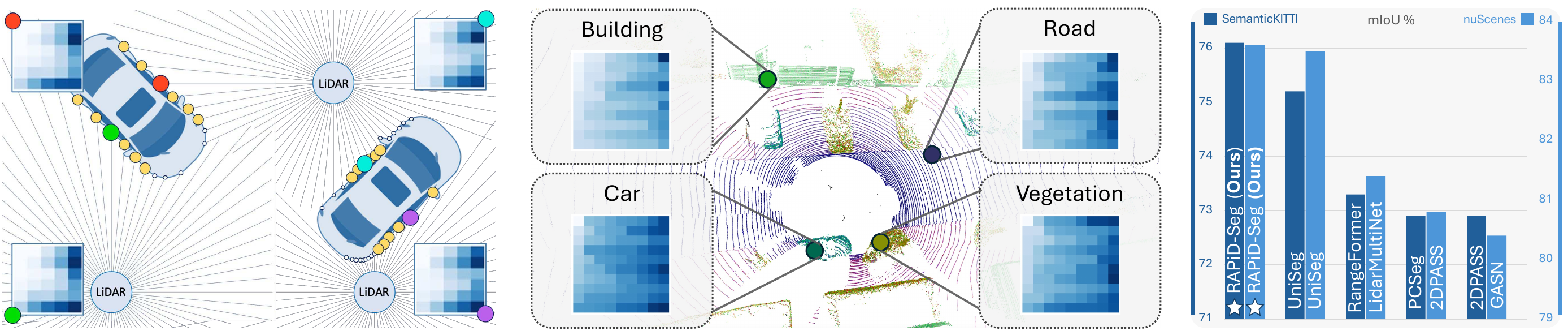}
    \vspace{-20pt}
    \captionof{figure}{\textls[-10]{\textbf{Left:} {\pddshort} exhibits excellent viewpoint invariance and geometric stability, visualizing comparable features around the vehicle door structure at varying ranges and viewpoints (feature matrix plots inset). \textbf{Middle:} {\pddshort} is distinctive in different semantic classes, as visualized by the matrices. Embedded {\pddshort} patterns corresponding to different points are visualized using a spectrum of colors, showcasing their capacity to represent different classes. \textbf{Right:} Our {\ourmodel} achieves superior results over SoTA methods on nuScenes~\cite{nuscenes2019} and  SemanticKITTI~\cite{behley2019semantickitti}.}}
    \label{fig:PDD_banner}
    \vspace{2.26pt}
\end{center}

\vspace{0.1cm}
\textls[-13]{Despite these advancements, a common challenge inherent to these methods is their poor isometric invariance, characterized by heavy reliance on solely coordinates and intensity data, which often results in suboptimal segmentation outcomes~\cite{li2023less,zhu2021cylindrical}; this is primarily due to poor translational invariance and visibility (\eg, occlusions), or sparse observations (\eg, at long range)~\cite{li2023memoryseg}, affecting data spatial distribution. }

\vspace{0.1cm}
In this work, we seek features that are (1) capable of capturing the localized geometric structure of neighboring points, (2) invariant to rotation and translation, and (3) applicable in noisy LiDAR outdoor environments. While numerous methods fulfill some of these requirements individually~\cite{jiang2018pointsift, liang2012geometric, melia2023rotationinvariant}, they fall short of addressing them all~\cite{flitton2013comparison}. %
Recognizing the need for higher-level features capable of capturing local geometry while potentially incorporating LiDAR point-specific attributes (\eg, intensity and reflectivity), we instead turn our attention to pointwise Distance Distribution (PDD) features~\cite{widdowson2022resolving,widdowson2023recognizing}. They are recognized for their remarkable efficacy in providing a robust and informative geometric representation of point clouds, excelling in both rotational and translational invariance while including intricate local geometry details.

\textls[-8]{However, directly employing PDD features is impractical due to their high-dimensional nature, and their significant memory and storage requirements for large-scale point clouds. PDD also overlooks local features, as including distant points subsequently dilutes the focus on immediate neighborhoods. }

\textls[-12]{As an enabler, we propose the \textbf{R}ange-\textbf{A}ware \textbf{P}o\textbf{i}ntwise \textbf{D}istance Distribution (\ourmodel) solution for LiDAR segmentation. As shown in~\cref{fig:PDD_banner}, it utilizes invariance of {\pddshort} features to rigid transformations and point cloud sparsity variations, while concentrating on compact features within specific local neighborhoods. Specifically, our method leverages inherent LiDAR isotropic radiation and semantic categorization for enhanced local representations while reducing computational burdens. Moreover, our formulation computes a 4D distance, incorporating both 3D geometric and reflectivity differences to enhance semantic segmentation fidelity. To compress high-dimensional {\pddshort} features into tractable voxel-wise embeddings, we propose a novel embedding approach with our class-aware double-nested AutoEncoder (AE) module. We further incorporate a channel-wise attention fusion and a two-stage training strategy to optimize the AE independently before integrating into the full network, enhancing performance and generalization capability.}

\textls[-5]{We conduct extensive experiments on SemanticKITTI~\cite{behley2019semantickitti} and nuScenes~\cite{nuscenes2019} datasets, upon which our approach surpasses the state-of-the-art (SoTA) segmentation performance.}

Overall, as shown in~\cref{fig:pipeline}, our contributions can be summarized as follows:
\textls[-10]{\begin{itemize}
\item[\textcolor{fig2g}{\raisebox{1pt}{\scriptsize$\blacksquare$}}] A novel {\pddfull} \textbf{\textsf{\textcolor{fig2g}{feature ({\pddshort})}}} for 3D LiDAR segmentation that ensures robustness to rigid transformations and viewpoints through isometry-invariant metrics within specific Regions of Interest, \ie, intra-ring (R-{\pddshort}) and intra-class (C-{\pddshort}).
\item[\textcolor{fig2b}{\raisebox{1pt}{\scriptsize$\blacksquare$}}] A novel method for \textbf{\textsf{\textcolor{fig2b}{embedding}}} {\pddshort} with class-aware double nested AE (RAPiD AE) to optimize the embedding of high-dimensional features, balancing efficiency and fidelity.
\item[\textcolor{fig2r}{\raisebox{1pt}{\scriptsize$\blacksquare$}}] A novel open-source network architecture \textbf{\textsf{\textcolor{fig2r}{{\ourmodel}}}}
~\footnote{The code is publicly available at: \url{https://github.com/l1997i/rapid_seg}.} 
and supporting training methodology, to enable modular LiDAR segmentation 
that achieves SoTA performance on SemanticKITTI (mIoU: \textbf{76.1}) and nuScenes (mIoU: \textbf{83.6}).

\begin{figure*}[thbp]
    \vspace{-5pt}
    \centering
    \includegraphics[width=1\linewidth]{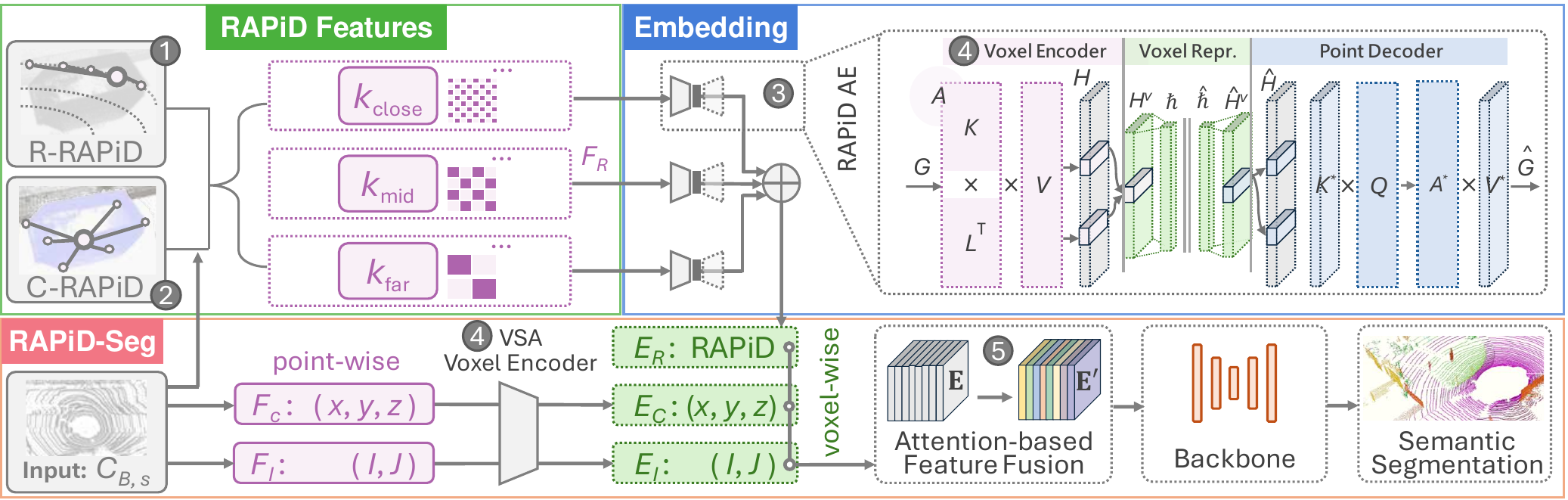}
    \vspace{-15pt} %
    \caption{\textls[-25]{Our proposed architecture for 3D segmentation framework leverages {\pddshort} features from the point cloud. We encode pointwise features into voxel-wise embeddings via the voxel encoder and multiple {\pddshort} AutoEncoders (RAPiD AE). After attention-based feature fusion, these fused embeddings go through the backbone network for segmentation results.}}
    \label{fig:pipeline}
    \vspace{-0.5cm}
\end{figure*}

\end{itemize}}

\vspace{-0.3cm}
\section{Related Work}
\vspace{-2pt}

\bdtitle{LiDAR-Based Semantic Segmentation}~\cite{af2s3net, genova2021learning, hou2022pointtovoxel, jiang2021guided, kong2023lasermix, kong2023rethinking, li2022self, li2023less, liong2020amvnet, liu2023segment, liu2023uniseg, rpvnet, tang2020searching, unal2022scribblesupervised, yan2021sparse, yan20222dpassb, ye2022efficient, ye2023lidarmultinet, zhu2021cylindrical, zhuang2021perception} is fundamental for LiDAR-driven scene perception, aiming to label each point in a point cloud sequence. The majority of the approaches~\cite{tang2020searching,zhu2021cylindrical,yan2021sparse,hou2022pointtovoxel,liu2023segment} solely rely on the point-based features of the point cloud, such as SPVCNN~\cite{tang2020searching} which introduces a point-to-voxel branch, using combined point-voxel features for segmentation. Cylinder3D~\cite{zhu2021cylindrical} proposes cylindrical partitioning with a UNet~\cite{cicek20163d} backbone variant. LiM3D~\cite{li2023less} utilizes coordinates combined with surface reflectivity attributes. Overall, such prior work relies solely on point-based features such as coordinates and intensity of the points, lacking an effective fusion mechanism, resulting in suboptimal performance~\cite{chen2022mppnet, li2023mseg3d, wang2022metarangeseg}. They are also susceptible to changes in viewpoint, distance, and point sparsity~\cite{li2023memoryseg} due to the lack of isometry across all inter-point distances. 

\bdtitle{Pointwise Distance Distribution (PDD)} captures the local context of each point in a unit cell by enumerating distances to neighboring points in order. It is an isometry invariant proposed by Widdowson \& Kurlin~\cite{widdowson2022resolving} to resolve the data ambiguity for periodic crystals, demonstrated through extensive pairwise comparisons across atomic 3D clouds from high-level periodic crystals of periodic structures~\cite{widdowson2022resolving,widdowson2022average,widdowson2023recognizing}. Though the effectiveness of PDD in periodic crystals and atomic clouds has been proved by the aforementioned studies, to date, no work has applied PDD features to outdoor 3D point clouds. 
In outdoor settings, common invariant features~\cite{liang2012geometric,jiang2018pointsift,melia2023rotationinvariant,li2022rotationinvariant} and Azimuth-normalized method~\cite{chen2022azinorm}
often face issues due to the irregular and sparse nature of the data, which is compounded by increased noise and environmental complexities~\cite{liang2012geometric,li2022rotationinvariant}. Furthermore, the computational demands make them less suitable for the vast scale of outdoor settings~\cite{jiang2018pointsift,melia2023rotationinvariant}. 
For instance, Melia~\etal~\cite{melia2023rotationinvariant} introduce a rotation-invariant feature that struggles with generalization across diverse point cloud densities and scales due to its computational cost and susceptibility to outdoor noise.
Recognizing these limitations, we identify an opportunity to leverage the PDD features in a new domain, where representing the local context of neighboring points in a transformation invariant and geometrically robust manner is crucial.
Drawing from the advantages of the PDD design, we propose the {\pddshort} feature, tailored specifically for LiDAR-based point clouds, to capture the localized geometry of neighboring structures.

\bdtitle{Point Cloud Representations and Embeddings} are fundamental for LiDAR-driven tasks. 3D point cloud can be represented either as point-based~\cite{qi2017pointneta,qi2017pointnetb,wu2022point,zhao2021point} or voxel-based~\cite{unal2022scribblesupervised,zhu2021cylindrical,hou2022pointtovoxel,liu2024u3ds3}, each serving different purposes rooted in their unique characteristics. Voxel-based representation involves dividing the 3D space into grids of voxels, which provides a structured way of handling sparse and irregular LiDAR data~\cite{zhu2021cylindrical,graham20183d,han2020occuseg,cicek20163d}. Previous work often utilizes Feedforward Neural Network (FNN)~\cite{unal2022scribblesupervised,zhu2021cylindrical,hou2022pointtovoxel} to facilitate the conversion of pointwise features into structured voxel-wise features by projecting them onto voxel grids and generating dense embeddings. The primary drawback of FNN conversion is the potential loss of fine-grained details inherent in the original point cloud~\cite{he2022voxela}. FNN may also struggle with the high dimensionality and sparsity of voxelized data, leading to computational inefficiency~\cite{shao2023ms23d}. Our method addresses these issues by partitioning the point cloud into voxel grids and applying a novel class-ware double nested AE with self-attention mechanisms~\cite{he2022voxela}. This approach not only retains more detailed information by focusing on local structures but also introduces high-dimensional feature compression, thereby improving computational efficiency.  %

\vspace{-0.2cm}
\section{Range-Aware Pointwise Distances (\pddshort)}
\label{sec:def_pdd}
\vspace{-3pt}

In the context of LiDAR segmentation, a principal challenge lies in the lack of robustness against rigid transformations such as rotation and translation, as well as against variations in viewpoint, points sparsity, and occlusion~\cite{li2023memoryseg,chen2023svqnet}. Traditional methods predominantly leverage 3D point coordinates to furnish spatial information~\cite{zhu2021cylindrical,hou2022pointtovoxel,unal2022scribblesupervised}; however, they may be inadequate in the scenes with poor visibility (\eg, occlusions) or sparse observations (\eg, at long range)~\cite{li2023memoryseg}. Such reliance on coordinates alone could lead to inaccuracies in recognizing object transformations or occlusions~\cite{li2023memoryseg}. Data augmentation, \eg, random geometric transformation can improve robustness under rigid transformations~\cite{zhu2021cylindrical,unal2022scribblesupervised}, but fail to guarantee comprehensive coverage of all potential transformations, resulting in vulnerability to previously unseen variations.

\subsection{Overview} 

To achieve a transformation-invariant 3D data representation, we observe that distances within rigid bodies (\eg, vehicles, roads, and buildings) remain constant under rigid transformations.
\textls[-12]{In light of this, we focus on assimilating the principle of the isometric invariant into the LiDAR-driven point cloud perception. Specifically, we delve into the PDD~\cite{widdowson2022resolving} — an isometry invariant that quantifies distances between adjacent points. For outdoor point clouds, vanilla PDD features are computationally intensive and susceptible to noise and sparsity. Our {\pddfull} ({\pddshort}), specifically designed for LiDAR data, instead calculates the PDD features for each point within specific Regions of Interest (RoI), which are typically associated with the intrinsic structure of LiDAR data.}

\bdtitle{Mathematical Formulation:} Given a fixed number $k>0$ representing the fixed number of point neighbors and a $u$-point cluster $P_\text{RoI}$ comprising no fewer than $k$ points based on RoI, the {\pddshort} is a $u \times k$ matrix, which retains both spatial distances and LiDAR reflectivity~\cite{li2023less} disparities between points. {\pddshort} are adapted to LiDAR sparsity at different distances by using range-specific parameters $k_{\text{close}}$, $k_{\text{mid}}$, $k_{\text{far}}$ for close, mid, and far ranges, respectively.

The $k$-point $\text{\pddshort}$ in region $P_\text{RoI}$ is defined as:
\vspace{-4pt}
\begin{equation}
\label{eq:rapid_geom}
\small {
\text{{\pddshort}}\left(P_\text{RoI}; k\right) = \text{sort}\left(\left[\ \text{sort}\left(\left[\ \boldsymbol{\rho}_{j, 1}, \ldots, \boldsymbol{\rho}_{j, k}\ \right]\right)\ \right]_{j=1}^u\right),
}
\vspace{-4pt}
\end{equation}
$\forall l \in\{1, \ldots, k\}, j \in\{1, \ldots, u\}$,\ \  $\bm{\rho}_{j, l}$ is given by:
\begin{equation}
\label{equ:spatial-reflect}
\bm{\rho}_{j, l}=\left\|\left[\begin{array}{c}
\mathvec{p}_j-\mathvec{p}_{j, l}, \ \ g\left(r_j\right)-g\left(r_{j, l}\right)
\end{array}\right]\right\|_2 ,
\end{equation}
where $\mathvec{p}_j$ and $\mathvec{p}_{j, l}$ denote the 3D coordinates of the $j$-th point and its $l$-th nearest neighbor within $P_\text{RoI}$, respectively; $r_j$ and $r_{j, l}$ represent the reflectivity values of $\mathvec{p}_j$ and $\mathvec{p}_{j, l}$, correspondingly; $\|\cdot\|_2$ is the Euclidean norm. $g: \mathbb{R} \rightarrow\left[D_{\min }, D_{\max }\right]$ is the reflectivity mapping function that maps the numerical range of reflectivity onto a consistent scale with the range of Euclidean distances between points:
\begin{align}
    g(r) &=\left(\frac{r-r_{\min }}{r_{\max }-r_{\min }}\right)\left(D_{\max }-D_{\min }\right)+D_{\min }, \label{equ:scale_reflec}
    \\
    D_{\min } &=\min _{j, l}\left\|\mathvec{p}_j-\mathvec{p}_{j, l}\right\|_2, \ \ D_{\max }=\max _{j, l}\left\|\mathvec{p}_j-\mathvec{p}_{j, l}\right\|_2. \label{equ:d_range}
\end{align}
where \( [D_{\min}, D_{\max}] \) is the range of the Euclidean norms of coordinate differences for all considered point pairs.

\enlargethispage{0.8cm}
The 4D distance in~\cref{equ:spatial-reflect} integrates material reflectivity~\cite{li2023less} into {\pddshort} features, enhancing feature representation and aiding in the discrimination of various materials and surfaces, which is crucial for accurate semantic segmentation.

\textls[-5]{For convenience to facilitate feature normalization and data alignment, we arrange {\pddshort} lexicographically by sorting~\cref{eq:rapid_geom}, where $\text{sort}(\cdot)$ on the inner and outer brackets sorts the elements $\bm{\rho}_{j, l}$ within each row $j$, and the sorted rows based on their first differing elements, both in ascending order~\cite{widdowson2022resolving}. It treats {\pddshort} matrices with the same geometric structure but different orders as identical.}

To improve the data robustness, we normalize $\mathvec{\rho}_{j,l}$ after excluding outliers exceeding a distance threshold $\delta$, substituting these entries with the maximum of the normalized distribution to represent significant inter-point distances.

\vspace{-0.4cm}
\subsection{Intra-Ring and Intra-Class {\pddshort}}
\label{sec:rc_pdd}
To enhance the {\pddshort} ability to capture local context, we propose {\rpdd} and {\cpdd}, based on the inherent structure of LiDAR data. 
The key benefit of {\cpdd} is its capacity to underscore the inherent traits of points within the same semantic class, achieved based on semantic labels before distance calculation.
As a complementary, {\rpdd} is versatile, regardless of semantic label availability.

\bdtitle{Intra-Ring {\pddshort} (\rpdd)} is a specialized variant of {\pddshort}, which confines the RoI to the ring encompassing the anchor points, thereby serving to economize on computational overhead while capitalizing on the inherent structural characteristics of LiDAR data (\cref{fig:pipeline}~\graynum{202}).

Given LiDAR sensors with fixed beam counts~\cite{nuscenes2019,behley2019semantickitti,sun2020scalabilitya,li2021durlara}, \eg, $ B \in \{32, 64, 128\} $, beams radiate isotropically around the vehicle at set angles. We segment the $m$-point LiDAR point cloud $C_{B, s}$ by beams, where $m = s \times B$, with $s$ representing measurements within a scan cycle and $B$ the number of laser beams. 
Suppose $\mathvec{p}_i$ with Cartesian coordinate $(x_i, y_i, z_i)$ is a point within $C_{B,s}$, its transformation to cylindrical coordinates $(\theta_i,\phi_i)$ is formulated as:
\begin{align} \small
    \theta_i &= \left\lfloor \arctan2(y_i, x_i) / \Delta \theta \right\rfloor, \\
    \phi_i &= \left\lfloor \arcsin\left[{z_i}\ (x_i^2+y_i^2+z_i^2)^{-1/2}\right] / \Delta \phi \right\rfloor,
\end{align}
where $ \Delta \theta $ and $ \Delta \phi $ denote the mean angular resolutions horizontally and vertically between adjacent beams, and $\arctan2(\cdot)$ is the 2-argument $\arctan(\cdot)$ function.

\textls[-20]{$C_{B, s}$ is then partitioned to $ B $ rings based on $ \phi_b$, \ie, $\small C_{B, s} = \bigcup{_{b=0}^{B-1}} R_b $, where each ring $\small R_b = \left\{ \left( \theta_{b,i}, \phi_{b,i} \right) \mid \phi_{b,i} = \phi_b, \ \forall i \leq m \right\}$; $ \bigcup $ signifies the cumulative union. %
{\rpdd} is thus defined as the cumulative concatenate $\oplus$ of {\pddshort} in each ring $R_b$,}
\vspace{-5pt}
\begin{equation}
    \text{{\rpdd}}(C_{B,s}; k) := \bigoplus{\ }_{b=0}^{B-1} {\ } \text{\pddshort}(R_b; k).
\end{equation}

\bdtitle{Intra-Class {\pddshort} (\cpdd)} concentrates on extracting point features within the confines of each semantic class (\cref{fig:pipeline}~\graynum{203}).
It is complementary to {\rpdd}, preserving the embedding fidelity within individual semantic class. 

Specifically, C-{\pddshort} concatenates {\pddshort} for each point $\mathvec{p}_j$ and its coresponding semantic label $y_j$ within semantic class $i$, across $N_c$ total classes:
\vspace{-3pt}
\begin{equation}
\begin{aligned}
    \text{{\cpdd}}(\{C_{B,s}, \mathcal{Y}\}; k) := \bigoplus{\ }_{i=0}^{N_c-1}{\ }\text{\pddshort}(S_i; k),
\end{aligned}
\end{equation}
\vspace{-0.1cm}
where $S_i = \{\mathvec{p}_j \mid y_j = i,\ \forall j \leq m\}$,\ \ $\mathcal{Y} = [y_j]_{j=1}^m$\ .

\vspace{-2pt}
\section{{\pddshort} Embedding}
\vspace{-2pt}
\label{sec:ring-wise-set-transformer}

The direct utilization of high-dimensional {\pddshort} features 
imposes a substantial computational burden. Efficient processing of large data volumes requires methods that can condense dimensionality while preserving data integrity~\cite{zhang2022visualizing}.

We explore data embedding methods to reduce computational costs while providing feature learning capacity. AE is efficient in embedding high-dimensional features utilizing the reconstruction loss~\cite{he2022masked, preechakul2022diffusion}. However, conventional AE, limited by their sensitivity to input order and reliance on fixed-size misinputs~\cite{kortvelesy2023permutationinvariant}, struggle to align with the unordered and variable size inherent in point clouds~\cite{he2022voxela}.

We propose a double nested AE structure with a novel class-aware embedding objective, using the Voxel-based Set Attention (VSA) module~\cite{he2022voxela} as a building block.
This facilitates superior contextual awareness and adaptability to unordered and variably-sized point cloud data. 

\begin{figure}
    \centering
    \includegraphics[width=0.75\linewidth]{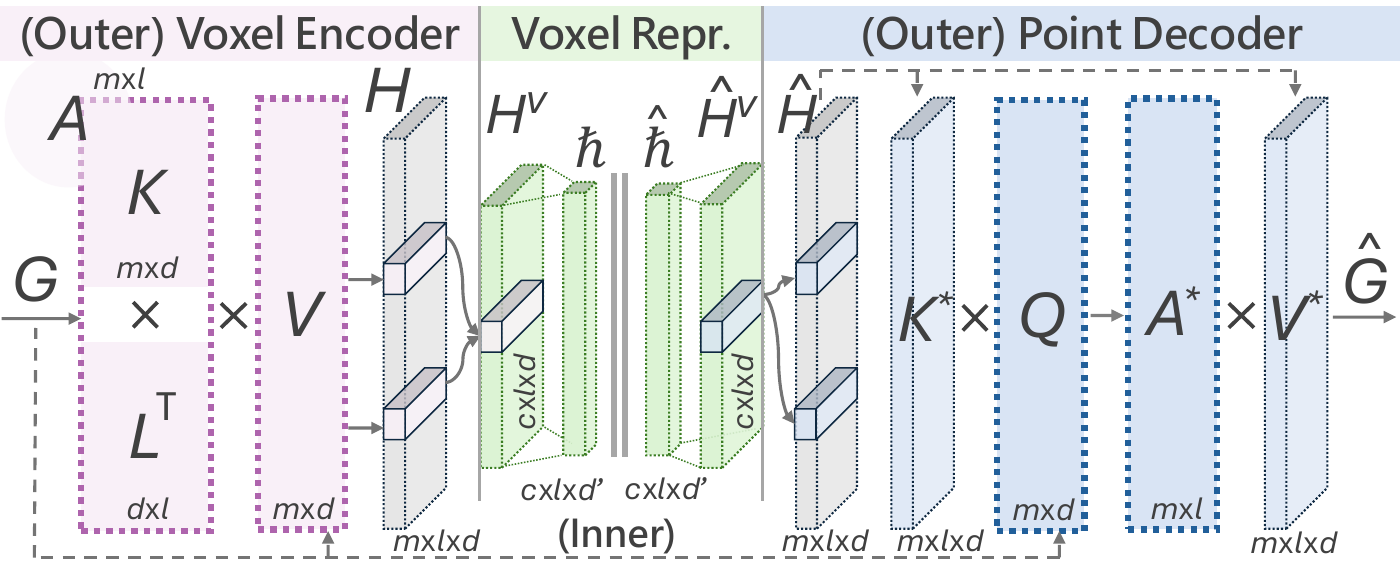}
    \vskip -6pt
    \caption{Our {\pddshort} AE consists of an Encoder, Convolution Layer, and Decoder module, aiming to reproduce the input features and generate the compressed voxel-wise {\pddshort} representation $\hbar$.}
    \label{fig:autoencoder_architechture}
    \vspace{-15pt}
\end{figure}

\bdtitle{Nested {\pddshort} AE},
illustrated in~\cref{fig:pipeline}~\graynum{204} and~\cref{fig:autoencoder_architechture}, is composed of two modules: the outer module is constituted by a VSA AE, primarily responsible for the point-to-voxel conversion. Within the inner module, which focuses on voxel-wise representation, we employ an additional AE specifically designed for dimensionality reduction. 
Specifically, considering an input of $m$-point {\pddshort} features $G \in \mathbb{R}^{m \times d}$, where each point encompasses $d$ features. We compress pointwise $G$ into a voxel-wise representation $ H^v \in \mathbb{R}^{c \times l \times d} $\ by an outer VSA Voxel Encoder (with $c$ reduced voxel-wise dimension and $l$ latent codes), then further reduces it to a compressed embedding $\hbar \in \mathbb{R}^{c \times l \times d'}$ with $d'$ features via an inner Encoder. The inner and outer VSA Point Decoders then reconstruct the output feature set $\hat{G}$ akin to $ G $.

\bdtitle{Inner Voxel-Wise Representation} is conducted via an inner AE, which takes $H^v $ from outer VSA AE as the input, yielding a lower-dimensional embedding $\hbar $ (dimension reduced from $d $ to $d' $). This process involves $u \times$ convolutional layers and a batch normalization layer~\cite{ioffe2015batch}: 
\begin{equation}
    \hbar^{(i)}=\text{BatchNorm} \circ \text{Conv}(\hbar^{(i-1)}),
\end{equation}
where $\hbar^{(0)} = H^v$, $ \hbar^{(u-1)} = \hbar$. The Convolutional Feed-Forward Network (ConvFFN) then promotes voxel-level information exchange, enhancing spatial feature interactivity. It maps reduced hidden features $\hbar$ to a 3D sparse tensor, indexed by voxel coordinates $X^v $, and employs dual depth-wise convolutions (DwConv) for spatial interactivity:
\begin{equation}
\hat \hbar = \text{DwConv}^{(2)} \circ \zeta \circ \text{DwConv}^{(1)} \left( \text{SpT}(\hbar, X^v) \right) ,
\end{equation}
where $ \zeta $ and $ \text{SpT}(\cdot) $ represents the non-linear activation and sparse tensor construction. Subsequently, $\hat{\hbar} \in \mathbb{R}^{c\times l \times d'} $ is reconstructed into $ \hat{H}^v \in \mathbb{R}^{c \times l \times d} $ through an inner Decoder, consisting of multiple DeConv Layers.

\bdtitle{Outer VSA AE} consists of a voxel encoder and point encoder. The voxel encoder project {\pddshort} features into key-value spaces, forming $K $ and $V $, followed by a cross-attention mechanism with a latent query $L $, producing an attention matrix $A $. The voxel-wise representation $H^v $ is the scatter sum~\cite{pytorchscatter} of pointwise $H$ to aggregate the value vectors $V $: 
\begin{align}
    H^v &= \text{Sum}_{\text{scatter}}(H, I^v), \quad H = \tilde{A}^\top V,    \\
    \tilde{A} &= \text{Softmax}_{\text{scatter}}(KL^\top, I^v), \quad (Q, V) = \text{Proj}(G),
    \label{equ:point_to_voxel}
\end{align}
where $\text{Proj}(\cdot)$ represents the linear projection, and $I^v$ is the voxel indices. 
The Point Decoder reconstructs the output set from the enriched hidden features $\hat{H}^v$. We start by broadcasting $\hat{H}^v$ based on $I^v$, resulting in $\hat{H} \in \mathbb{R}^{m\times l \times d}$. The output $\hat{G}$ is analogous to the operational paradigm of Voxel Encoder:
\begin{align}
\hat{G} &= \tilde{A^*}^\top V^*, \quad \tilde{A^*} = \text{Softmax}({A^*}), \\
A^* &= \left[ K^*_i Q_i^\top \right]_{i=1}^m, \quad (K^*, V^*) = \text{Proj}(\mathcal{\hat{H}}).
\end{align}

\bdtitle{Class-Aware Embedding Objective} addresses the issue of non-uniqueness in embeddings produced by the AE, where various distinct inputs yield approximately the same embedding, leading to inaccurate representation~\cite{badger2023depth,kumar2020implicit}. Our objective aims to facilitate the generation of AE embeddings that demonstrate robust semantic class discriminability.

Specifically, we introduce a novel class-aware contrastive loss $\mathcal{L}_\text{contr}$ in~\cref{equ:contr_loss}, which aims to maximize the distance between embeddings of different semantic classes while minimizing the distance of the same class.
{
\vspace{-3pt}
\small
\begin{align}
    \mathcal{L}_\text{contr} &= \frac{1}{m} \sum_{i=1}^{m} \left[ {\frac{1}{|P(i)|}} \sum_{p \in P(i)} {\text{ReLU}(\alpha_p - \text{sim}(H_i, H_p))} \right. \nonumber \\
    &\quad\quad + \left. {\frac{1}{|N(i)|}} \sum_{n \in N(i)} {\text{ReLU}(\text{sim}(H_i, H_n) - \alpha_n)} \right], \label{equ:contr_loss} \\
    \mathcal{L}_\text{recon} &= \frac{1}{m \cdot d} \sum_{i=1}^{m} \sum_{j=1}^d (G_{i,j} - \hat{G}_{i,j})^2, \label{equ:recon_loss}
    \end{align}
\normalsize 
\textls[-5]{where $i$ represents the point index; $P(i)$ and $N(i)$ are the indices of the nearest point in the same (\underline{P}ositive) and different class (\underline{N}egative) relative to point $i$, respectively; $\text{sim}(\cdot)$ is a similarity measure; $\alpha_p$ and $\alpha_n$ is the positive and negative margin controlling separation between classes, respectively. The pointwise representations $H_i$, $H_p$, and $H_n$ are broadcasted from $\hbar$ based on voxel indices $I^v$.}
}

We further combine this with the MSE~\cite{marmolin1986subjective} reconstruction loss in~\cref{equ:recon_loss}. The overall loss function, with $\lambda$ balancing the fidelity of reconstruction with the distinctiveness of the embeddings, is formulated as: $\mathcal{L}_\text{total} = \mathcal{L}_\text{recon} + \lambda \mathcal{L}_\text{contr}$.

\vspace{-8pt}
\section{{\ourmodel} for 3D LiDAR Segmentation}
\label{sec:rapid_seg}
\vspace{-2pt}

\textls[-10]{We present {\ourmodel}, a 3D LiDAR segmentation network leveraging {\pddshort} features. It incorporates multiple complementary features via the channel-wise fusion mechanism~\cite{hu2018squeezeandexcitation}. {\ourmodel} takes point cloud as input and performs single-modal (LiDAR-only) 3D semantic segmentation. Specifically, the input point cloud has three types of point-wise features: (a) \textbf{coordinates-based features} $F_C = \{ p_{_C} \mid p_{_C}=(x,y,z) \}$; (b) \textbf{intensity-based features} $F_I = \{ p_{_I} \mid p_{_I}=(I, J) \}$, where $I$ and $J$ are intensity and reflectivity; (c) \textbf{RAPiD features} $F_R = \{ p_{_R} \mid p_{_R}=\text{RAPiD}(P_\text{RoI};k) \}$. The RoI varies depending on whether it is R- or C-RAPiD. }

As shown in~\cref{fig:pipeline}, $F_C \oplus F_I$ and $F_R$ are fed to a VSA voxel encoder~\graynum{205} and RAPiD AE~\graynum{204} for voxelization, resulting in voxel-wise representation $E_C$, $E_I$ and $E_R$. Subsequently, we incorporate the complementary voxel-wise features via the channel-wise fusion mechanism (\cref{sec:fusion_with_attention}). The backbone net takes in them for LiDAR segmentation prediction. Additionally, we introduce two effective {\ourmodel} variants to expedite the convergence of the extensive segmentation network towards an optimal solution. 

\vspace{-3pt}
\subsection{{\pddshort} Channel-Wise Fusion with Attention}
\label{sec:fusion_with_attention}
\vspace{-3pt}
To combine various LiDAR point attributes, a highly efficient method for feature fusion is still a research topic~\cite{li2023less,unal2022scribblesupervised}. Current methods, which typically concatenate different LiDAR attributes (\eg, intensity~\cite{nunes2023temporal,ye2022lidarmultinet,liu2023segment}, reflectivity~\cite{li2023less}, PLS~\cite{unal2022scribblesupervised}) face issues with increased dimensionality and complexity, and the need for balancing weights to avoid biased training towards dominant attributes.

\textls[-25]{We thus propose a novel {\pddshort} Fusion with Attention module (FuAtten), pioneering application of channel-wise attention mechanism~\cite{hu2018squeezeandexcitation} in LiDAR point attribute fusion.
With FuAtten, we effectively fuse $E_C$, $E_I$ and $E_R$, emphasizing informative features adaptively while suppressing less relevant ones across different channels.}

\begin{figure}[tp]
    \centering
    \includegraphics[width=0.6\linewidth]{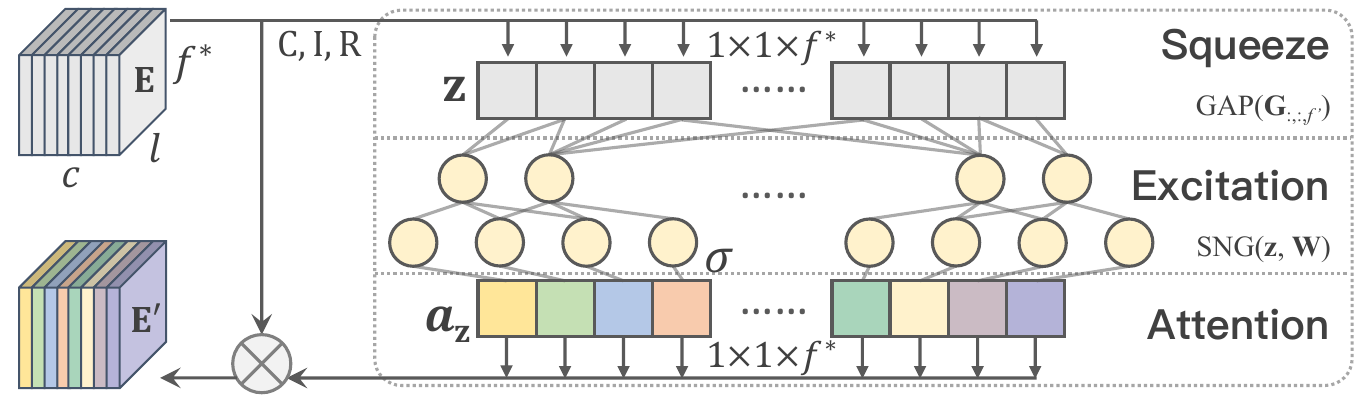}
    \caption{Our feature fusion module with channel-wise attention.}
    \vspace{-0.5cm}
    \label{fig:chann_atten}
\end{figure}

\textls[-10]{As shown in~\cref{fig:pipeline}~\graynum{206} and~\cref{fig:chann_atten}, we initially concatenate $E_C$, $E_I$ and $E_R$ along the voxel dimension, resulting in $\mathbf{E}$. In the squeeze operation, Global Average Pooling (GAP)~\cite{lin2014network} condenses each channel $f'$ of the tensor into a representative channel descriptor. These descriptors form a channel-level embedding $\mathbf{z} = [z_{f'}]_{{f'=1}}^{f^*}$:}
\begin{equation}
\small
\mathbf{z} = \left[\text{GAP}(\mathbf{E}_{:,:,f'})\right]_{{f'=1}}^{f^*} = \frac{1}{c \times l} \left[\sum_{c'=0}^{c-1} \sum_{l'=0}^{l-1} \mathbf{E}_{c',l',f'}\right]_{{f'=1}}^{f^*},
\end{equation}
where $f^*$ is the dimension of the concatenated feature.
A channel-wise attention $\mathvec{a}_\mathbf{z} \in \mathbb{R}^{1 \times 1 \times f^*}$ is then computed through the excitation step: $\mathvec{a}_\mathbf{z} = \sigma\left(\mathbf{W}_{2} \ \delta(\mathbf{W}_1 \mathbf{z}\right))$,
where $\delta$ and $\sigma$ represent the ReLU and Sigmoid activation.
With each element in $\mathvec{a}_\mathbf{z}$ reflects the attention allocated to the corresponding feature of the voxel, the channel-wise fused features $\mathbf{E}' = \mathvec{a}_\mathbf{z} \cdot \mathbf{E}$.

\subsection{{\ourmodel} Architectures for 3D Segmentation}

The complexity inherent in 3D LiDAR-driven networks typically exacerbates the end-to-end training process~\cite{hou2022pointtovoxel,unal2022scribblesupervised,li2023less}, since the extensive parameter space increases the propensity for overfitting, slow convergence, and the possibility of settling into local minima~\cite{li2023less}.

\vspace{-10pt}
\begin{figure}
    \centering
    \includegraphics[width=0.85\linewidth]{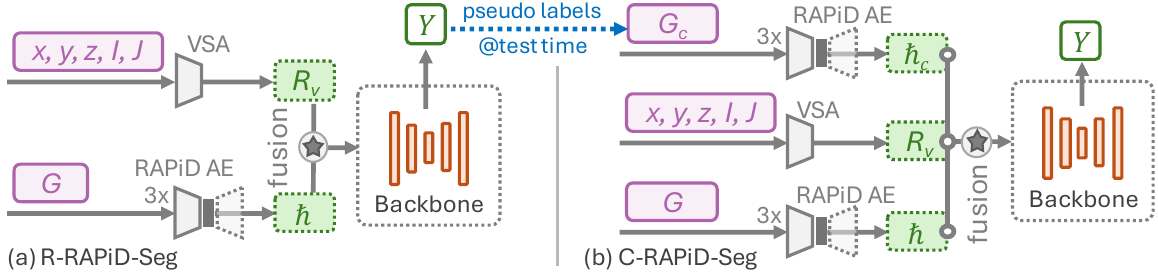}
    \caption{Two variants of {\ourmodel}. (a) \textbf{R-{\ourmodel}} utilizes R-{\pddshort} features, and (b) \textbf{C-{\ourmodel}} utilizes both R- and C-{\pddshort} features for better performance.}
    \vspace{-8pt}
    \label{fig:varient_framework}
\end{figure}

As shown in~\cref{fig:varient_framework}, we propose two novel and effective variants of {\ourmodel} for quicker and better performance. R-{\ourmodel} involves AE training from scratch, followed by C-{\ourmodel} utilizing C-{\pddshort} features with the pre-trained AE and backbone for enhanced performance.

\bdtitle{R-{\ourmodel}.} 
We first construct the lightweight R-{\ourmodel} (\cref{fig:varient_framework}~(a)) for fast 3D segmentation. R-{\ourmodel} adopts an early fusion scheme utilizing LiDAR original points and R-{\pddshort} features. Specifically, 3D coordinate $F_C$ and intensity-based features $F_I$ are voxelized to voxel-wise representations $R_{v}=E_C \oplus E_I$ based on VSA voxelization~\cite{he2022voxela}. We get the compressed $d'$-dimension voxel-wise R-{\pddshort} representations $\hbar$ (\cref{sec:ring-wise-set-transformer}) from RAPiD AE. $\hbar$ is then fused into the voxel-wise representations $R_{v*} = \text{FuAtten}(R_{v}, \hbar)$, where $\text{FuAtten}(\cdot)$ is Fusion with Attention in~\cref{sec:fusion_with_attention}. The fused voxel-wise features are taken into the backbone network for 3D segmentation.

\bdtitle{C-{\ourmodel}.} 
To facilitate the embedding fidelity within individual semantic classes, we also design a class-aware framework with C-{\pddshort} features, \ie, C-{\ourmodel} (\cref{fig:varient_framework}~(b)). Specifically, we fuse the voxel-wise C-{\pddshort} $\hbar_C$ into the representations $R_{v*} = \text{FuAtten}(R_{v}, \hbar \oplus \hbar_C)$. C-{\pddshort} requires class labels to compute the features regarding the semantic categories. Since the ground-truth labels are missing during the test time, we generate the reliable pseudo labels $\tilde{\mathcal{Y}}$ with a pretrained R-{\ourmodel} based on confidence (\cref{fig:varient_framework}~(b) \textcolor{bluearrow}{blue dotted arrow}). Subsequently, the pseudo voxel-wise C-{\pddshort} $\hbar_C$ are generated through $\tilde{\mathcal{Y}}$. We fuse them into the voxel-wise representations for 3D segmentation with the bonebone network.

\vspace{-12pt}
\section{Evaluation}
\vspace{-2pt}
\label{sec:experiments}

Following the popular practice of LiDAR segmentation methods~\cite{zhu2021cylindrical,hou2022pointtovoxel,liu2023uniseg}, we evaluate our proposed {\ourmodel} network against SoTA 3D LiDAR segmentation approaches on the SemanticKITTI~\cite{behley2019semantickitti} and nuScenes~\cite{nuscenes2019} datasets.

\vspace{-12pt}
\subsection{Experimental Setup}
\vspace{-4pt}

\bdtitle{Datasets:} 
SemanticKITTI~\cite{behley2019semantickitti} comprises 22 point cloud sequences, with sequences \texttt{00}-\texttt{10}, \texttt{08}, and \texttt{11}-\texttt{21} for training, validation, and testing. 19 classes are chosen for training and evaluation by merging classes with similar motion statuses and discarding sparsely represented ones. Meanwhile, nuScenes~\cite{nuscenes2019} has 1000 driving scenes; 850 are for training and validation, with the remaining 150 for testing. 16 classes are used for LiDAR semantic segmentation, following the amalgamation of akin classes and the removal of rare ones.

\bdtitle{Evaluation Protocol:} Following the popular practice of~\cite{behley2019semantickitti,zhu2021cylindrical,liu2023pcseg}, we adopt the Intersection-over-Union (IoU) of each class and mean IoU~(mIoU) of all classes as the evaluation metric. The IoU of class $i$ is defined as $\text{IoU}_{i} = \text{TP}_{i} / (\text{TP}_{i}+\text{FP}_{i}+\text{FN}_{i})$, where $\text{TP}_{i}$, $\text{FP}_{i}$ and $\text{FN}_{i}$ denote the true positive, false positive and false negative of class $i$, respectively. 

\bdtitle{Implementation Details:} We construct the point-voxel backbone based on the Minkowski-UNet34~\cite{choy20194d} (re-implemented by PCSeg~\cite{liu2023pcseg} codebase), which is the open-access backbone with SoTA results to date. Before the AE training stage, we first generate pointwise {\pddshort} features (\cref{sec:def_pdd}), which yield three {\pddshort} outputs for each frame, corresponding to different $k$ values based on range. Notably, this stage does not impose additional burdens on the subsequent overall training. The number of maximum training epochs for AE and the whole network is set as 100 and the initial learning rate is set as $10^{-3}$ with SGD optimizer. We use 2 epochs to warm up the network and adopt the cosine learning rate schedule for the remaining epochs. All experiments are conducted on $4\times$ NVIDIA A100 GPUs ($1\times$ for inference).

\vspace{-6pt}
\subsection{Experimental Results}
\vspace{-2pt}

In~\cref{tab:semkitti_testset} and~\cref{tab:nuscenes_testset}, we showcase the performance of our {\ourmodel} LiDAR segmentation method on SemanticKITTI and nuScenes {\testset} in comparison with published leading contemporary SoTA approaches to demonstrate its superior efficacy (\textbf{top 10} results shown; full rankings in Supplementary Material).

\begin{table*}[ht]
\caption{{Quantitative results of {\ourmodel} and top-10 SoTA segmentation methods on SemanticKITTI}~\cite{behley2019semantickitti} \testset; \textbf{Best}/\underline{2nd best} highlighted; $\blacktriangle$ for multi-modal methods.}
\vskip -0.3cm
\label{tab:semkitti_testset}
\centering
\setlength{\tabcolsep}{1.5pt}
\renewcommand{\arraystretch}{0.95} %
\begin{adjustbox}{width=\textwidth}
\begin{tabular}{@{}r|c|c|c|c|c|c|c|c|c|c|c|c|c|c|c|c|c|c|c|c@{}}
\topr
Method & mIoU & \rotatebox{0}{car} & \rotatebox{0}{bicy} & \rotatebox{0}{moto} & \rotatebox{0}{truc} & \rotatebox{0}{o.veh} & \rotatebox{0}{ped} & \rotatebox{0}{b.list} & \rotatebox{0}{m.list} & \rotatebox{0}{road} & \rotatebox{0}{park} & \rotatebox{0}{walk} & \rotatebox{0}{o.gro} & \rotatebox{0}{build} & \rotatebox{0}{fenc} & \rotatebox{0}{veg} & \rotatebox{0}{trun} & \rotatebox{0}{terr} & \rotatebox{0}{pole} & \rotatebox{0}{sign}
\\
\topr
Cylinder3D~\cite{zhu2021cylindrical} & 68.9 & 97.1 & 67.6 & 63.8 & 50.8 & 58.5 & 73.7 & 69.2 & 48.0 & 92.2 & 65.0 & 77.0 & 32.3 & 90.7 & 66.5 & 85.6 & 72.5 & 69.8 & 62.4 & 66.2
\\
AF2S3Net~\cite{af2s3net} & 69.7 & 94.5 & 65.4 & \textbf{86.8} & 39.2 & 41.1 & \textbf{80.7} & 80.4 & \underline{74.3} & 91.3 & 68.8 & 72.5 & \textbf{53.5} & 87.9 & 63.2 & 70.2 & 68.5 & 53.7 & 61.5 & \underline{71.0}
\\
RPVNet~\cite{rpvnet} & 70.3 & 97.6 & 68.4 & 68.7 & 44.2 & 61.1 & 75.9 & 74.4 & 73.4 & \textbf{93.4} & 70.3 & \textbf{80.7} & 33.3 & \underline{93.5} & 72.1 & 86.5 & \underline{75.1} & 71.7 & 64.8 & 61.4
\\
SDSeg3D~\cite{li2022self} & 70.4  & 97.4 & 58.7 & 54.2 & 54.9 & 65.2 & 70.2 & 74.4 & 52.2 & 90.9 & 69.4 & 76.7 &  41.9 & 93.2 & 71.1 & 86.1 & 74.3 & 71.1 & 65.4 & 70.6
\\
GASN~\cite{ye2022efficient} & 70.7 & 96.9 & 65.8 &  58.0 & 59.3 & 61.0 & \underline{80.4} & \textbf{82.7} & 46.3 & 89.8 & 66.2 & 74.6 & 30.1 & 92.3 & 69.6 & \underline{87.3} & 73.0 & 72.5 & 66.1 & \textbf{71.6}
\\
PVKD~\cite{hou2022pointtovoxel} & 71.2 & 97.0 & 67.9 & 69.3 & 53.5 & 60.2 & 75.1 & 73.5 & 50.5 & 91.8 & 70.9 & 77.5 & 41.0 & 92.4 & 69.4 & 86.5 & 73.8 & 71.9 & 64.9 & 65.8
\\
2DPASS~\cite{yan20222dpassb} & 72.9 & 97.0 & 63.6 & 63.4 & 61.1 & 61.5 & 77.9 & \underline{81.3} & 74.1 & 89.7 & 67.4 & 74.7 & 40.0 & \underline{93.5} & \textbf{72.9} & 86.2 & 73.9 & 71.0 & 65.0 & 70.4
\\
PCSeg~\cite{liu2023pcseg} & 72.9 & 97.5 & 51.2 & 67.6 & 58.6 & 68.6 & 78.3 & 80.9 & \textbf{75.6} & 92.5 & 71.5 & 78.3 & 36.9 & 93.1 & 71.4 & 85.4 & 73.6 & 69.9 & 66.1 & 68.7
\\
RangeFormer~\cite{kong2023rethinking} & 73.3 & 96.7 & 69.4 & 73.7 & 59.9 & 66.2 & 78.1 & 75.9 & 58.1 & 92.4 & 73.0 & 78.8 & 42.4 & 92.3 & 70.1 & 86.6 & 73.3 & \underline{72.8} & \underline{66.4} & 66.6
\\
$\blacktriangle$UniSeg~\cite{liu2023uniseg} & \underline{75.2} & \textbf{97.9} & \textbf{71.9} & 75.2 & \underline{63.6} & \underline{74.1} & 78.9 & 74.8 & 60.6 & \underline{92.6} & \underline{74.0} & \underline{79.5} & \underline{46.1} & 93.4 & \underline{72.7} & \textbf{87.5} & \textbf{76.3} & \textbf{73.1} & \textbf{68.3} & 68.5 \\
\midr
\rowcolor{dodgerblue!15}\textbf{\ourmodel~(Ours)} & \textbf{76.1} & \underline{97.7} & \underline{71.1} & \underline{76.2} & \textbf{72.5} & \textbf{80.7} & 79.9 & 79.1 & 59.8 & 91.8 & \textbf{78.2} & 78.6 & 46.0 & \textbf{93.6} & 72.1 & 86.9 & 74.6 & 72.3 & 65.9 & 68.5 \\
\bottomr
\end{tabular}
\end{adjustbox}
\vspace{-5pt}
\end{table*}
\begin{table*}[ht]
\caption{Quantitative results of {\ourmodel} and top-10 SoTA segmentation methods on nuScenes~\cite{nuscenes2019} \testset; \textbf{Best}/\underline{2nd best} highlighted; $\blacktriangle$ for multi-modal methods.}
\vskip -0.3cm
\label{tab:nuscenes_testset}
\centering
\setlength{\tabcolsep}{2pt} %
\renewcommand{\arraystretch}{0.95} %
\scriptsize %
\begin{adjustbox}{width=\textwidth}
\begin{tabular}{@{}r|c|c|c|c|c|c|c|c|c|c|c|c|c|c|c|c|c@{}}
\topr
Method & mIoU & \rotatebox{0}{barr} &  \rotatebox{0}{bicy} & \rotatebox{0}{bus} & \rotatebox{0}{car} & \rotatebox{0}{const} & \rotatebox{0}{motor} & \rotatebox{0}{ped} & \rotatebox{0}{cone} & \rotatebox{0}{trail} & \rotatebox{0}{truck} & \rotatebox{0}{driv} & \rotatebox{0}{other} &
\rotatebox{0}{walk} & \rotatebox{0}{terr} & \rotatebox{0}{made} & \rotatebox{0}{veg} \\
\toprr
$\blacktriangle$PMF~\cite{zhuang2021perception}
&77.0 & 82.0& 40.0& 81.0& 88.0& 64.0 &79.0& 80.0 & 76.0& 81.0& 67.0& 97.0& 68.0& 78.0& 74.0& 90.0& 88.0\\
Cylinder3D~\cite{zhu2021cylindrical} & 77.2 & 82.8 & 29.8 & 84.3 & 89.4 & 63.0 & 79.3 & 77.2 & 73.4 & 84.6 & 69.1 & \underline{97.7} & 70.2 & 80.3 & 75.5 & 90.4 & 87.6  \\

AMVNet~\cite{liong2020amvnet} &77.3 &  80.6 & 32.0 & 81.7 & 88.9 & 67.1 & 84.3 & 76.1 & 73.5 & 84.9 & 67.3 & 97.5 & 67.4 & 79.4 & 75.5 & 91.5 & 88.7\\

SPVCNN~\cite{tang2020searching} &77.4& 80.0& 30.0 & 91.9 & 90.8 & 64.7 & 79.0 & 75.6 & 70.9 & 81.0 & 74.6 & 97.4 & 69.2 & 80.0 & 76.1 & 89.3 & 87.1 \\
AF2S3Net~\cite{af2s3net} &78.3 & 78.9 & 52.2 & 89.9 & 84.2 & 77.4 & 74.3 & 77.3 & 72.0 & 83.9 & 73.8 & 97.1 & 66.5 & 77.5 & 74.0 & 87.7 & 86.8 \\
$\blacktriangle$2D3DNet~\cite{genova2021learning} & 80.0 & 83.0 & 59.4 & 88.0 &85.1 & 63.7 & 84.4 & 82.0 & 76.0 & 84.8 & 71.9 & 96.9 & 67.4 & 79.8 & 76.0 & \underline{92.1} & 89.2 \\
GASN~\cite{ye2022efficient} &80.4 & \underline{85.5} & 43.2 & 90.5 & \underline{92.1} & 64.7 & 86.0 & \underline{83.0} & 73.3 & 83.9 & 75.8 & 97.0 & \underline{71.0} & \underline{81.0} & \bf{77.7} & 91.6 & \bf{90.2}\\
2DPASS~\cite{yan20222dpassb} & 80.8 & 81.7 & 55.3 & 92.0 & 91.8 & 73.3 & 86.5 & 78.5 & 72.5 & 84.7 & 75.5 & 97.6 & 69.1 & 79.9 & 75.5 & 90.2 & 88.0\\
$\blacktriangle$LidarMultiNet~\cite{ye2023lidarmultinet} & 81.4 & 80.4 & 48.4 & \underline{94.3} & 90.0 & 71.5 & 87.2 & \bf{85.2} & \bf{80.4} & \underline{86.9} & 74.8 & \bf{97.8} & 67.3 & 80.7 & 76.5 & \underline{92.1} & \underline{89.6} \\
$\blacktriangle$UniSeg~\cite{liu2023uniseg} & \underline{83.5} & \bf{85.9} & \bf{71.2} & 92.1 & 91.6 & \underline{80.5} & \bf{88.0} & 80.9 & 76.0 & 86.3 & \underline{76.7} & \underline{97.7} & \bf{71.8} & 80.7 & \underline{76.7} &  91.3 & 88.8 \\
\midr
\rowcolor{dodgerblue!15}\textbf{\ourmodel~(Ours)} & \bf{83.6} & 84.8 & \underline{64.3} & \textbf{95.0} & \textbf{92.2} & \textbf{84.6} & \underline{87.9} & 81.8 & \underline{76.8} & \textbf{88.5} & \textbf{79.0} & \textbf{97.8} & 66.6 & \textbf{81.2} & \underline{76.7} &  \textbf{92.5} & 88.4
\\\bottomr
\end{tabular}
\end{adjustbox}
\vspace{-12pt}
\end{table*}

\textls[-6]{Our method significantly outperforms others, especially for rigid object categories (\eg, \textls[-20]{\texttt{truck}, \texttt{o.veh}, \texttt{park}}, \etc), primarily due to our fusion of both localized 3D geometry and material reflectivity within {\pddshort} features, which enable segmentation based on material properties and local rigid structures. Remarkably, our single-modal methodology outperforms multi-modal approaches~\cite{liu2023uniseg,yan20222dpassb,ye2023lidarmultinet,liong2020amvnet,rpvnet,kong2023rethinking}, suggesting superior efficacy of our {\pddshort} features over alternative modalities like RGB and range images.}
Our inference time (105ms per frame) is comparable to other contemporary approaches~\cite{liu2023uniseg,yan20222dpassb,zhu2021cylindrical}.
Furthermore, we present supporting qualitative results in~\cref{fig:visual_results}. Whereas the baseline method struggles with accurate vehicle type differentiation, ours achieves consistent segmentation (more visualization results in the supplementary materials).

\begin{table}[htbp]
\vspace{-0.5cm}
    \centering
    \begin{minipage}{0.47\linewidth}

\centering
\setlength{\tabcolsep}{1pt} %
\renewcommand{\arraystretch}{0.95} %
\begin{adjustbox}{width=\linewidth}
{
\small{
\begin{tabular}{@{}c|c|c|c|cc@{}}
\topr
\multicolumn{3}{c|}{\textbf{RAPiD Features}} & \multicolumn{1}{c|}{\multirow{2}{*}{Attention}} & \multirow{2}{*}{mIoU} & \multirow{2}{*}{$\Delta$}\\
Geometric & Reflectivity & Embedding & \multicolumn{1}{c|}{} & & \\ 
\midr
 & & & & 70.04 & \tabequ{Baseline} \\
   &  &  &  $\checkmark$  & 70.46 & \tabup{+0.42} \\
$\checkmark$ &         &         &   & 71.21 & \tabup{+1.17}\\
$\checkmark$ & $\checkmark$ &         &  & 71.93 & \tabup{+1.89}\\
$\checkmark$ &         $\checkmark$ &         $\checkmark$&  & 72.15 & \tabup{+2.11}\\ 
 &         $\checkmark$ & $\checkmark$ &  $\checkmark$ & 71.80 & \tabup{+1.76}\\
  $\checkmark$ & & $\checkmark$ & $\checkmark$ & 72.32 & \tabup{+2.28}\\
 $\checkmark$ & $\checkmark$ & & $\checkmark$& 72.78 & \tabup{+2.74}\\
 \midr
\rowcolor{dodgerblue!15}$\checkmark$ & $\checkmark$ & $\checkmark$ & $\checkmark$ & \textbf{73.02} & \tabup{+2.98}\\
\bottomr
\end{tabular}
}}
\end{adjustbox}
\vspace{0.2cm}
\caption{\textls[-35]{Component-wise ablation of {\ourmodel} on the SemanticKITTI {\validset}: Geometric (\cref{eq:rapid_geom}), Reflectivity (\cref{equ:scale_reflec}), (RAPiD) Embedding (\cref{sec:ring-wise-set-transformer}), (channel) Attention (\cref{sec:fusion_with_attention}).}}
\label{tab:ablations}
\end{minipage}
\vspace{\compactlen}

    \hspace{0.2cm}
    \begin{minipage}{.48\linewidth}
\vspace{-6pt}
    \centering
    \includegraphics[width=\linewidth]{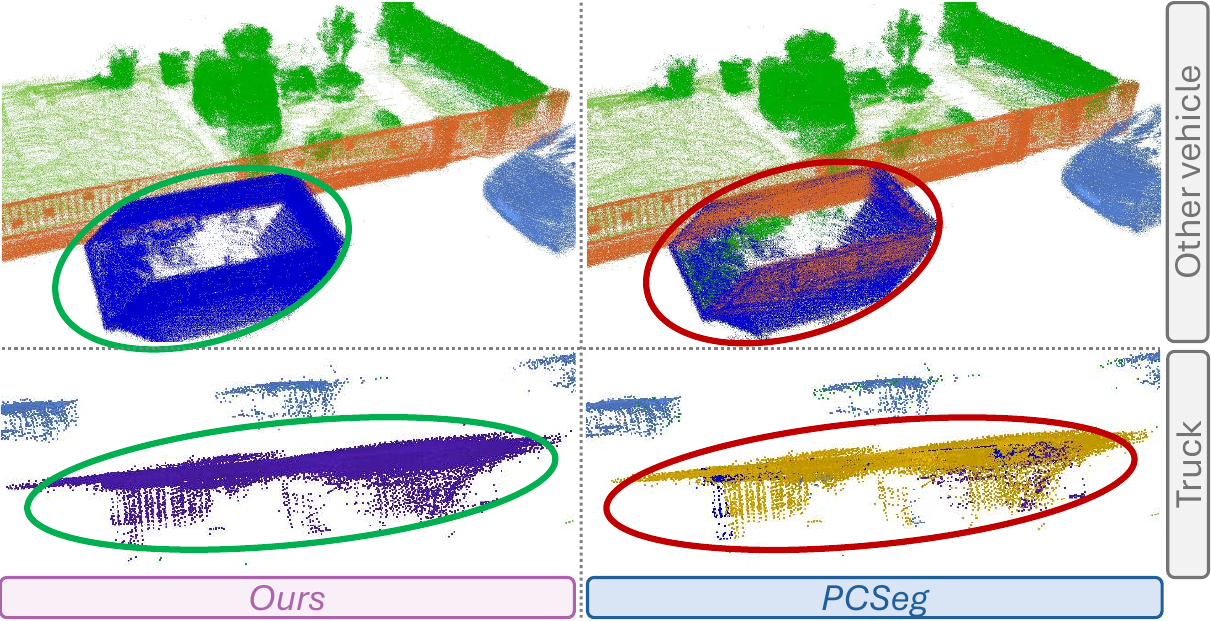}
    \vspace{6pt}
    \captionof{figure}{\textls[-20]{Comparing our results and PCSeg (baseline) under multi-scan visualization, showing improved segmentation results.}}
    \label{fig:visual_results}
\end{minipage}
    \vspace{-1cm}
\end{table}
\vspace{-0.3cm}

\vspace{-6pt}
\subsection{Ablation Studies}
\vspace{-4pt}

\bdtitle{Effectiveness of Components:} In~\cref{tab:ablations}, we ablate each component of {\ourmodel} step by step and report the performance on the SemanticKITTI {\validset}. We start from a baseline model, achieving an mIoU of 70.04 on {\validset}. Incorporating the {\pddshort} features leads to a notable increase in mIoU (+1.17), underscoring the efficacy of {\pddshort} in enhancing segmentation performance. Building upon the {\pddshort} framework, the integration of reflectivity further elevates our mIoU to 71.93, with an overall improvement of +1.89 from the baseline. This significant gain illustrates the critical role of reflectivity in capturing different object materials. Utilizing {\pddshort} AE to get the voxel-wise representation, we observe an additional improvement, taking the mIoU to 72.15 (+2.11 compared to the baseline), which demonstrates its capability in processing feature representations. We subsequently performed a module-by-module reduction to analyze the model performance. Specifically, without RAPiD-related modules (Geometric/Reflectivity/Embedding), mIoU drops from 73.02 to 70.46 - attributable to the lack of RAPiD features, which are robust to transformation and noise. Our ablation studies show that while disabling individual components leads to some improvement over the baseline, the peak performance is only attained when all components function together. This highlights the indispensable contribution of each component in maximizing segmentation accuracy.

\begin{table}[tp]
\caption{Effects of {\pddshort} and Reflectivity features compared to other configurations on SemanticKITTI \validset. PDD: the compressed PDD embeddings; {\pddshort}-R: only {\pddshort}, without reflectivity feature fusion; {\pddshort}+R: both {\pddshort} and reflectivity.}
\label{tab:effect_rap_reflec}
\renewcommand{\arraystretch}{0.95} %
\setlength{\tabcolsep}{13pt} %
\centering
\vspace{-0.3cm}
\begin{adjustbox}{width=1\linewidth}
{\small{
\begin{tabular}{@{}r|c|ccccccc@{}}
\toprule
 Conf.& mIoU & truck&  o.veh&park &walk &o.gro &build &fence\\
\toprule
  BaseL~\cite{liu2023pcseg}& 70.0& 59.8& \underline{70.3}&69.2& 76.9&\underline{36.2}& \underline{93.7}&69.6\\
  PDD~\cite{widdowson2022resolving}& 66.2& 40.3&  65.8&67.5& 74.6&33.1& 92.8&68.2\\
 {\pddshort}-R&         \underline{71.8}&         \underline{62.5}&    69.0&\underline{70.3}& \underline{77.4}&35.8& 93.6&\underline{70.8}\\
 \midrule
 \rowcolor{dodgerblue!15} {\pddshort}+R& \textbf{73.0}& \textbf{70.4}& \textbf{78.5}& \textbf{75.8}& \textbf{78.9}& \textbf{44.2}& \textbf{94.2} & \textbf{73.1}\\
 \bottomrule
\end{tabular}
}}
\end{adjustbox}
\vspace{-0.4cm}
\end{table}

\bdtitle{\textls[-10]{Effectiveness of {\pddshort} and Reflectivity Features:}} In~\cref{tab:effect_rap_reflec}, we validate the efficacy of the proposed {\pddshort} features. The direct application of PDD results in a substantial mIoU decrease of 3.8, with most category IoUs falling below those of the baseline method. This indicates vanilla PDD features are not well-suited for LiDAR segmentation. While our proposed {\pddshort} (without reflectivity disparities in~\cref{equ:spatial-reflect}) demonstrates an improvement of 1.4 in mIoU over the baseline. Moreover, the integration of reflectivity with {\pddshort} features significantly enhances performance, yielding 2.6 mIoU increase over the baseline. Notably, our approach exhibits superior performance in the segmentation of most rigid object categories compared to other configurations. 
\cref{tab:hyper_k} further shows the {\pddshort} outperforms PDD in varying $k$ values at different ranges on SemanticKITTI {\validset}, with non-uniform $k$ configurations yielding the most significant improvements in mIoU for both datasets, highlighting the effectiveness of the range-aware design of {\pddshort}.

\begin{table}[htbp]
\vspace{-0.1cm}
    \centering
    \begin{minipage}{.48\linewidth}
\vspace{-0.9cm}
\caption{3D segmentation results of different variants of {\ourmodel} (ours) on SemanticKITTI {\validset}.}
\label{tab:variant_framework}
\centering
\vspace{-0.18cm}
\setlength{\tabcolsep}{5pt} %
\renewcommand{\arraystretch}{1.2} %
\begin{adjustbox}{width=1\linewidth}
{\small{
\begin{tabular}{@{}r|l|cccl@{}}
\topr
Method& mIoU \% & car&  ped&o.gro &pole\\
\toprr
Baseline& 70.0 & 97.2& 78.1& 35.4&63.5 \\
R-{\ourmodel}& 72.3 \tabup{+2.3}& 97.4& 77.4& \textbf{45.0}&62.4\\
C-{\ourmodel}&         \textbf{73.0}~\tabup{+3.0}&         \textbf{97.7}&    \textbf{79.3}& 44.6&\textbf{66.4}\\
 \bottomr
\end{tabular}
}}
\end{adjustbox}
\vspace{-1cm}
\end{minipage}
    \hspace{0.2cm}
    \begin{minipage}{.47\linewidth}
\caption{Effects of $k$ at various ranges.}
\label{tab:hyper_k}
\centering
\vspace{-0.18cm}
\setlength{\tabcolsep}{2pt}
\renewcommand{\arraystretch}{0.98} %
\begin{adjustbox}{width=1\linewidth}
{\small{
\begin{tabular}{@{}l|cccl|cccl@{}}
\topr
  &\multicolumn{4}{c|}{PDD~\cite{widdowson2023recognizing}}&\multicolumn{4}{c}{{\pddshort} (ours)}\\
  &$k_\text{near}$& $k_\text{mid}$& $k_\text{far}$& mIoU&$k_\text{near}$& $k_\text{mid}$& $k_\text{far}$&mIoU\\
\toprr
  \multirow{3}{*}{\rotatebox[origin=c]{90}{SemK}} &7& 7& 7& 64.74 \tabdown{-7.1}&7& 7& 7&72.04 \tabup{+0.2}\\
                     &5& 5& 5& 65.18 \tabdown{-6.6}&5& 5& 5&72.28 \tabup{+0.5}\\
                     &10& 7& 5& 66.23 \tabdown{-5.6}& \textbf{10}& \textbf{7}& \textbf{5}& \textbf{73.02} \tabup{+1.2} \\
\toprr
  \multirow{3}{*}{\rotatebox[origin=c]{90}{nuScene}} & 6& 6& 6& 72.19 \tabdown{-6.5} & 6& 6& 6& 78.76 \tabup{+0.1} \\
                     & 3& 3& 3& 73.68 \tabdown{-5.0} & 3& 3& 3& 79.43 \tabup{+0.8} \\
                     &8& 6& 3& 72.24 \tabdown{-6.4} & \textbf{8}& \textbf{6}& \textbf{3}& \textbf{79.91} \tabup{+1.3} \\
\bottomr
\end{tabular}
}}
\end{adjustbox}
\vspace{0.6\compactlen}
\end{minipage}

    \vspace{-0.3cm}
\end{table}

\bdtitle{Effectiveness of {\pddshort} AE and Architecture Variants:} In \cref{tab:ablations}, we replace our {\pddshort} AE with a commonly-used ConvFNN~\cite{unal2022scribblesupervised,zhu2021cylindrical,hou2022pointtovoxel} to generate voxel-wise embeddings. Our AE demonstrated a 0.24 mIoU improvement over the FNN. 
In~\cref{tab:variant_framework}, we investigate the impact of the variant architectures, \ie, R-{\ourmodel} and C-{\ourmodel}. Utilizing R-{\pddshort} features, R-{\ourmodel} yields an mIoU of 72.3, which outperforms the baseline method by +2.3 mIoU. Concurrent fusing of both R- and C-{\ourmodel} features in R-{\ourmodel} improves performance by +0.7 mIoU, which shows a +3.0 overall mIoU enhancement.

\bdtitle{Effectiveness of Feature Fusion with Channel-wise Attention:}
In \cref{tab:ablations}, we assess the prevalent approach of direct feature concatenation. Our method, employing feature fusion with channel-wise attention, enhances mIoU by 0.87, thereby conclusively demonstrating its effectiveness.

\bdtitle{Effectiveness of Backbone Networks:} In~\cref{tab:backbones}, we evaluate the performance across multiple backbone networks to demonstrate the versatility and ease of integration of our modules. The results highlight the adaptability and effectiveness of our approach on both pointwise (P) and Voxel-wise (V) backbone architectures. Specifically, the point-based PTv2 backbone achieves an mIoU of 72.6, showing strong performance, particularly in the \textls[-50]{\texttt{car}} and \textls[-50]{\texttt{other ground}} categories. Among the voxel-based methods, Minkowski-UNet stands out with the highest overall mIoU of 73.0, excelling in the \textls[-50]{\texttt{pedestrian}} and \textls[-50]{\texttt{pole}} categories. 

\begin{table}[thp]
\vspace{-0.2cm}
\caption{Effects of using different backbones on SemanticKITTI \validset, where P and V for Point- and Voxel-wsie methods.}
\label{tab:backbones}
\vspace{-0.2cm}
\centering
\setlength{\tabcolsep}{11pt} %
\renewcommand{\arraystretch}{0.95} %
\begin{adjustbox}{width=0.98\linewidth}
{\small{
\begin{tabular}{@{}c|r|l|cccl@{}}
\topr
 Repr.&Backbone& mIoU \% & car&  ped&o.gro &pole\\
\toprr
  P&PTv2 ~\cite{wu2022point}& 72.6 \tabup{+2.8}& \textbf{97.4}& 77.4& \textbf{45.0}&62.4\\
  V&Cylinder3D~\cite{zhu2021cylindrical}& 69.8 & 96.9&  74.2& 37.9&63.0\\
 \rowcolor{dodgerblue!15} V&Minkowski-UNet~\cite{choy20194d}&         \textbf{73.0}~\tabup{+3.2}&         97.2&    \textbf{78.1}& 44.2&\textbf{65.9}\\
 \bottomr
\end{tabular}
}}
\end{adjustbox}
\vspace{-0.5cm}
\end{table}

\section{Conclusion and Discussion}
\vspace{-0.1cm}
\label{sec:conclusion}

This paper presents a novel \textbf{R}ange-\textbf{A}ware \textbf{P}o\textbf{i}ntwise \textbf{D}istance Distribution ({\pddshort}) feature and the {\ourmodel} network for LiDAR segmentation, adeptly overcoming the constraints of single-modal LiDAR methods. The rigid transformation invariance and enhanced focus on local details of {\pddshort} significantly boost segmentation accuracy. {\ourmodel} integrates a two-stage training approach with reflectivity-guided 4D distance metrics and a class-aware nested AE, achieving SoTA results on the SemanticKITTI and nuScenes datasets. Notably, our single-modal method surpasses the performance of multi-modal methods, indicating the superior efficacy of {\pddshort} features even compared to other modalities, including RGB and range images.

\textls[-20]{Our {\pddshort} features hold significant potential for application in various tasks such as object detection~\cite{li2024traildet}, point cloud registration, semi-/weakly- supervised learning, and extend to promising applications in multi-modal research in the future.}

\clearpage  %

\hypersetup{colorlinks=true, urlcolor=black}
\bibliographystyle{splncs04}
\bibliography{abbr,ref}

\begin{thebibliography}{10}
\providecommand{\url}[1]{\texttt{#1}}
\providecommand{\urlprefix}{URL }
\providecommand{\doi}[1]{https://doi.org/#1}

\bibitem{badger2023depth}
Badger, B.L.: Depth and {{Representation}} in {{Vision Models}} (2023)

\bibitem{behley2019semantickitti}
Behley, J., Garbade, M., Milioto, A., Quenzel, J., Behnke, S., Stachniss, C.,
  Gall, J.: {{SemanticKITTI}}: {{A Dataset}} for {{Semantic Scene
  Understanding}} of {{LiDAR Sequences}}. In: Int. Conf. Comput. Vis. pp.
  9296--9306 (2019)

\bibitem{nuscenes2019}
Caesar, H., Bankiti, V., Lang, A.H., Vora, S., Liong, V.E., Xu, Q., Krishnan,
  A., Pan, Y., Baldan, G., Beijbom, O.: {Nuscenes: {{A}} Multimodal Dataset for
  Autonomous Driving}. In: IEEE Conf. Comput. Vis. Pattern Recog. pp.
  11618--11628 (2020)

\bibitem{chen2022azinorm}
Chen, S., Wang, X., Cheng, T., Zhang, W., Zhang, Q., Huang, C., Liu, W.:
  {AziNorm}: {{Exploiting}} the {{Radial Symmetry}} of {{Point Cloud}} for
  {{Azimuth-Normalized 3D Perception}}. In: IEEE Conf. Comput. Vis. Pattern
  Recog. pp. 6387--6396 (2022)

\bibitem{chen2023svqnet}
Chen, X., Xu, S., Zou, X., Cao, T., Yeung, D.Y., Fang, L.: {{SVQNet}}: {{Sparse
  Voxel-Adjacent Query Network}} for {{4D Spatio-Temporal LiDAR Semantic
  Segmentation}}. In: Int. Conf. Comput. Vis. pp. 8569--8578 (2023)

\bibitem{chen2022mppnet}
Chen, X., Shi, S., Zhu, B., Cheung, K.C., Xu, H., Li, H.: {{MPPNet}}:
  {{Multi-frame Feature Intertwining}} with~{{Proxy Points}} for~{{3D Temporal
  Object Detection}}. In: Avidan, S., Brostow, G., Ciss{\'e}, M., Farinella,
  G.M., Hassner, T. (eds.) Eur. Conf. Comput. Vis. pp. 680--697. {Springer
  Nature Switzerland} (2022)

\bibitem{af2s3net}
Cheng, R., Razani, R., Taghavi, E., Li, E., Liu, B.: ({{AF}})2-{{S3Net}}:
  {{Attentive}} feature fusion with adaptive feature selection for sparse
  semantic segmentation network. In: IEEE Conf. Comput. Vis. Pattern Recog. pp.
  12547--12556 (2021)

\bibitem{choy20194d}
Choy, C., Gwak, J., Savarese, S.: {{4D Spatio-Temporal ConvNets}}: {{Minkowski
  Convolutional Neural Networks}}. In: IEEE Conf. Comput. Vis. Pattern Recog.
  pp. 3070--3079 (2019)

\bibitem{cicek20163d}
{\c C}i{\c c}ek, {\"O}., Abdulkadir, A., Lienkamp, S.S., Brox, T., Ronneberger,
  O.: {{3D U-Net}}: {{Learning Dense Volumetric Segmentation}} from {{Sparse
  Annotation}}. In: Ourselin, S., Joskowicz, L., Sabuncu, M.R., Unal, G.,
  Wells, W. (eds.) Med. Image Comput. Comput. Assist. Interv. pp. 424--432.
  {Springer International Publishing} (2016)

\bibitem{flitton2013comparison}
Flitton, G., Breckon, T.P., Megherbi, N.: {A Comparison of {{3D}} Interest
  Point Descriptors with Application to Airport Baggage Object Detection in
  Complex {{CT}} Imagery}. Pattern Recognition  \textbf{46}(9),  2420--2436
  (2013)

\bibitem{Geiger2012}
Geiger, A., Lenz, P., Urtasun, R.: {Are We Ready for Autonomous Driving? The
  {{KITTI}} Vision Benchmark Suite}. In: IEEE Conf. Comput. Vis. Pattern Recog.
  pp. 3354--3361 (2012)

\bibitem{genova2021learning}
Genova, K., Yin, X., Kundu, A., Pantofaru, C., Cole, F., Sud, A., Brewington,
  B., Shucker, B., Funkhouser, T.: Learning {{3D Semantic Segmentation}} with
  only {{2D Image Supervision}}. In: Int. Conf. 3D Vis. pp. 361--372 (2021)

\bibitem{graham20183d}
Graham, B., Engelcke, M., {van der Maaten}, L.: {{3D Semantic Segmentation}}
  with {{Submanifold Sparse Convolutional Networks}}. In: IEEE Conf. Comput.
  Vis. Pattern Recog. (2018)

\bibitem{han2020occuseg}
Han, L., Zheng, T., Xu, L., Fang, L.: {{OccuSeg}}: {{Occupancy-Aware 3D
  Instance Segmentation}}. In: IEEE Conf. Comput. Vis. Pattern Recog. pp.
  2940--2949 (2020)

\bibitem{he2022voxela}
He, C., Li, R., Li, S., Zhang, L.: Voxel {{Set Transformer}}: {{A Set-to-Set
  Approach}} to {{3D Object Detection}} from {{Point Clouds}}. In: IEEE Conf.
  Comput. Vis. Pattern Recog. pp. 8407--8417. {IEEE} (2022)

\bibitem{he2022masked}
He, K., Chen, X., Xie, S., Li, Y., Doll{\'a}r, P., Girshick, R.: Masked
  {{Autoencoders Are Scalable Vision Learners}}. In: IEEE Conf. Comput. Vis.
  Pattern Recog. pp. 16000--16009 (2022)

\bibitem{hou2022pointtovoxel}
Hou, Y., Zhu, X., Ma, Y., Loy, C.C., Li, Y.: Point-to-{{Voxel Knowledge
  Distillation}} for {{LiDAR Semantic Segmentation}}. In: IEEE Conf. Comput.
  Vis. Pattern Recog. (2022)

\bibitem{hu2018squeezeandexcitation}
Hu, J., Shen, L., Sun, G.: Squeeze-and-{{Excitation Networks}}. In: IEEE Conf.
  Comput. Vis. Pattern Recog. (2018)

\bibitem{ioffe2015batch}
Ioffe, S., Szegedy, C.: Batch {{Normalization}}: {{Accelerating Deep Network
  Training}} by {{Reducing Internal Covariate Shift}}. In: Int. {{Conf}}.
  {{Mach}}. {{Learn}}. pp. 448--456 (2015)

\bibitem{jiang2021guided}
Jiang, L., Shi, S., Tian, Z., Lai, X., Liu, S., Fu, C.W., Jia, J.: Guided
  {{Point Contrastive Learning}} for {{Semi-supervised Point Cloud Semantic
  Segmentation}}. In: Int. Conf. Comput. Vis. (2021)

\bibitem{jiang2018pointsift}
Jiang, M., Wu, Y., Zhao, T., Zhao, Z., Lu, C.: {{PointSIFT}}: {{A SIFT-like
  Network Module}} for {{3D Point Cloud Semantic Segmentation}} (2018)

\bibitem{kong2023rethinking}
Kong, L., Liu, Y., Chen, R., Ma, Y., Zhu, X., Li, Y., Hou, Y., Qiao, Y., Liu,
  Z.: Rethinking {{Range View Representation}} for {{LiDAR Segmentation}}. In:
  Int. Conf. Comput. Vis. (2023)

\bibitem{kong2023lasermix}
Kong, L., Ren, J., Pan, L., Liu, Z.: {{LaserMix}} for {{Semi-Supervised LiDAR
  Semantic Segmentation}}. In: IEEE Conf. Comput. Vis. Pattern Recog. pp.
  21705--21715 (2023)

\bibitem{kortvelesy2023permutationinvariant}
Kortvelesy, R., Morad, S., Prorok, A.: Permutation-{{Invariant Set
  Autoencoders}} with {{Fixed-Size Embeddings}} for {{Multi-Agent Learning}}.
  In: Int. {{Conf}}. {{Auton}}. {{Agents Multiagent Syst}}. pp. 1661--1669.
  {International Foundation for Autonomous Agents and Multiagent Systems}
  (2023)

\bibitem{kumar2020implicit}
Kumar, A., Poole, B.: On {{Implicit Regularization}} in $\beta$-{{VAEs}}. In:
  Int. {{Conf}}. {{Mach}}. {{Learn}}. pp. 5480--5490 (2020)

\bibitem{li2023memoryseg}
Li, E., Casas, S., Urtasun, R.: {{MemorySeg}}: {{Online LiDAR Semantic
  Segmentation}} with a {{Latent Memory}}. In: Int. Conf. Comput. Vis. pp.
  745--754 (2023)

\bibitem{li2022self}
Li, J., Dai, H., Ding, Y.: {Self-Distillation for Robust {{LiDAR}} Semantic
  Segmentation in Autonomous Driving}. In: Eur. Conf. Comput. Vis. pp. 659--676
  (2022)

\bibitem{li2023mseg3d}
Li, J., Dai, H., Han, H., Ding, Y.: {{MSeg3D}}: {{Multi-Modal 3D Semantic
  Segmentation}} for {{Autonomous Driving}}. In: IEEE Conf. Comput. Vis.
  Pattern Recog. pp. 21694--21704 (2023)

\bibitem{li2021durlara}
Li, L., Ismail, K.N., Shum, H.P.H., Breckon, T.P.: {{DurLAR}}: {{A
  High-Fidelity}} 128-{{Channel LiDAR Dataset}} with {{Panoramic Ambient}} and
  {{Reflectivity Imagery}} for {{Multi-Modal Autonomous Driving Applications}}.
  In: Int. Conf. 3D Vis. pp. 1227--1237 (2021)

\bibitem{li2024traildet}
Li, L., Qiao, T., Shum, H.P.H., Breckon, T.P.: {TraIL-Det:
  Transformation-Invariant Local Feature Networks for 3D LiDAR Object Detection
  with Unsupervised Pre-Training}. In: Brit. Mach. Vis. Conf. (2024)

\bibitem{li2023less}
Li, L., Shum, H.P.H., Breckon, T.P.: Less is {{More}}: {{Reducing Task}} and
  {{Model Complexity}} for {{Semi-Supervised 3D Point Cloud Semantic
  Segmentation}}. In: IEEE Conf. Comput. Vis. Pattern Recog. (2023)

\bibitem{li2022rotationinvariant}
Li, X., Li, R., Chen, G., Fu, C.W., {Cohen-Or}, D., Heng, P.A.: A
  {{Rotation-Invariant Framework}} for {{Deep Point Cloud Analysis}}. IEEE
  Trans. Vis. Comput. Graph.  \textbf{28}(12),  4503--4514 (2022)

\bibitem{liang2012geometric}
Liang, J., Lai, R., Wong, T.W., Zhao, H.: {Geometric Understanding of Point
  Clouds Using {{Laplace-Beltrami}} Operator}. In: IEEE Conf. Comput. Vis.
  Pattern Recog. pp. 214--221 (2012)

\bibitem{lin2014network}
Lin, M., Chen, Q., Yan, S.: Network {{In Network}}. In: Int. Conf. Learn.
  Represent. (2014)

\bibitem{liong2020amvnet}
Liong, V.E., Nguyen, T.N.T., Widjaja, S., Sharma, D., Chong, Z.J.: {{AMVNet}}:
  {{Assertion-based Multi-View Fusion Network}} for {{LiDAR Semantic
  Segmentation}} (2020)

\bibitem{liu2024u3ds3}
Liu, J., Yu, Z., Breckon, T.P., Shum, H.P.H.: {{U3DS3}}: {{Unsupervised}} {{3D
  Semantic Scene Segmentation}}. In: IEEE Winter Conf. Appl. Comput. Vis. pp.
  3759--3768 (2024)

\bibitem{liu2023pcseg}
Liu, Y., Bai, Y., Chen, R., Hou, Y., Shi, B., Li, Y., Kong, L.: {{PCSeg}}: {{An
  Open Source Point Cloud Segmentation Codebase}}.
  \url{https://github.com/PJLab-ADG/PCSeg} (2023)

\bibitem{liu2023uniseg}
Liu, Y., Chen, R., Li, X., Kong, L., Yang, Y., Xia, Z., Bai, Y., Zhu, X., Ma,
  Y., Li, Y., Qiao, Y., Hou, Y.: {{UniSeg}}: {{A Unified Multi-Modal LiDAR
  Segmentation Network}} and the {{OpenPCSeg Codebase}}. In: Int. Conf. Comput.
  Vis. pp. 21662--21673 (2023)

\bibitem{liu2023segment}
Liu, Y., Kong, L., Cen, J., Chen, R., Zhang, W., Pan, L., Chen, K., Liu, Z.:
  {Segment Any Point Cloud Sequences by Distilling Vision Foundation Models}.
  In: Adv. Neural Inform. Process. Syst. (2023)

\bibitem{marmolin1986subjective}
Marmolin, H.: Subjective {{MSE}} measures. IEEE Trans. Syst. Man Cybern.
  \textbf{16}(3),  486--489 (1986)

\bibitem{melia2023rotationinvariant}
Melia, O., Jonas, E., Willett, R.: Rotation-{{Invariant Random Features
  Provide}} a {{Strong Baseline}} for {{Machine Learning}} on {{3D Point
  Clouds}}. Trans. Mach. Learn. Res.  (2023)

\bibitem{nunes2023temporal}
Nunes, L., Wiesmann, L., Marcuzzi, R., Chen, X., Behley, J., Stachniss, C.:
  Temporal {{Consistent 3D LiDAR Representation Learning}} for {{Semantic
  Perception}} in {{Autonomous Driving}}. In: IEEE Conf. Comput. Vis. Pattern
  Recog. (2023)

\bibitem{preechakul2022diffusion}
Preechakul, K., Chatthee, N., Wizadwongsa, S., Suwajanakorn, S.: Diffusion
  {{Autoencoders}}: {{Toward}} a {{Meaningful}} and {{Decodable
  Representation}}. In: IEEE Conf. Comput. Vis. Pattern Recog. pp. 10619--10629
  (2022)

\bibitem{pytorchscatter}
PyTorch: Scatter: {{PyTorch Scatter Documentation}}.
  \url{https://pytorch-scatter.readthedocs.io/en/latest/functions/scatter.html}
  (2023)

\bibitem{qi2017pointnetb}
Qi, C.R., Su, H., Mo, K., Guibas, L.J.: {{PointNet}}: {{Deep Learning}} on
  {{Point Sets}} for {{3D Classification}} and {{Segmentation}}. In: IEEE Conf.
  Comput. Vis. Pattern Recog. pp. 652--660 (2017)

\bibitem{qi2017pointneta}
Qi, C.R., Yi, L., Su, H., Guibas, L.J.: {{PointNet}}++: {{Deep Hierarchical
  Feature Learning}} on {{Point Sets}} in a {{Metric Space}}. In: Adv. Neural
  Inform. Process. Syst. vol.~30 (2017)

\bibitem{shao2023ms23d}
Shao, Y., Tan, A., Yan, T., Sun, Z., Zhang, Y., Liu, J.: {{MS23D}}: {{A 3D
  Object Detection Method Using Multi-Scale Semantic Feature Points}} to
  {{Construct 3D Feature Layer}} (2023)

\bibitem{sun2024empirical}
Sun, J., Xu, X., Kong, L., Liu, Y., Li, L., Zhu, C., Zhang, J., Xiao, Z., Chen,
  R., Wang, T., et~al.: {An Empirical Study of Training State-of-the-Art LiDAR
  Segmentation Models}. arXiv:2405.14870  (2024)

\bibitem{sun2020scalabilitya}
Sun, P., Kretzschmar, H., Dotiwalla, X., Chouard, A., Patnaik, V., Tsui, P.,
  Guo, J., Zhou, Y., Chai, Y., Caine, B., Vasudevan, V., Han, W., Ngiam, J.,
  Zhao, H., Timofeev, A., Ettinger, S., Krivokon, M., Gao, A., Joshi, A.,
  Zhang, Y., Shlens, J., Chen, Z., Anguelov, D.: Scalability in {{Perception}}
  for {{Autonomous Driving}}: {{Waymo Open Dataset}}. In: IEEE Conf. Comput.
  Vis. Pattern Recog. pp. 2446--2454 (2020)

\bibitem{tang2020searching}
Tang, H., Liu, Z., Zhao, S., Lin, Y., Lin, J., Wang, H., Han, S.: Searching
  {{Efficient 3D Architectures}} with {{Sparse Point-Voxel Convolution}}. In:
  Eur. Conf. Comput. Vis. (2020)

\bibitem{unal2022scribblesupervised}
Unal, O., Dai, D., Van~Gool, L.: {Scribble-Supervised {{LiDAR}} Semantic
  Segmentation}. In: IEEE Conf. Comput. Vis. Pattern Recog. (2022)

\bibitem{wang2022metarangeseg}
Wang, S., Zhu, J., Zhang, R.: {Meta-{{RangeSeg}}: {{LiDAR}} Sequence Semantic
  Segmentation Using Multiple Feature Aggregation}. IEEE Robot. Autom. Lett.
  \textbf{7}(4),  9739--9746 (2022)

\bibitem{widdowson2022resolving}
Widdowson, D., Kurlin, V.: {Resolving the Data Ambiguity for Periodic
  Crystals}. In: Adv. Neural Inform. Process. Syst. vol.~35, pp. 24625--24638
  (2022)

\bibitem{widdowson2023recognizing}
Widdowson, D., Kurlin, V.: Recognizing {{Rigid Patterns}} of {{Unlabeled Point
  Clouds}} by {{Complete}} and {{Continuous Isometry Invariants With No False
  Negatives}} and {{No False Positives}}. In: IEEE Conf. Comput. Vis. Pattern
  Recog. pp. 1275--1284 (2023)

\bibitem{widdowson2022average}
Widdowson, D., Mosca, M.M., Pulido, A., Cooper, A.I., Kurlin, V.: {Average
  Minimum Distances of Periodic Point Sets \textendash{} Foundational
  Invariants for Mapping Periodic Crystals}. MATCH Commun. Math. Comput. Chem.
  \textbf{87}(3),  529--559 (2022)

\bibitem{wu2022point}
Wu, X., Lao, Y., Jiang, L., Liu, X., Zhao, H.: {Point Transformer {{V2}}:
  {{Grouped}} Vector Attention and Partition-Based Pooling}. In: Adv. Neural
  Inform. Process. Syst. (2022)

\bibitem{rpvnet}
Xu, J., Zhang, R., Dou, J., Zhu, Y., Sun, J., Pu, S.: {{{RPVNet}}: {{A}} Deep
  and Efficient Range-Point-Voxel Fusion Network for {{LiDAR}} Point Cloud
  Segmentation}. In: Int. Conf. Comput. Vis. pp. 16024--16033 (2021)

\bibitem{yan2021sparse}
Yan, X., Gao, J., Li, J., Zhang, R., Li, Z., Huang, R., Cui, S.: {Sparse Single
  Sweep {{LiDAR}} Point Cloud Segmentation via Learning Contextual Shape Priors
  from Scene Completion}. In: Conf. AAAI Artif. Intell. vol.~35, pp. 3101--3109
  (2021)

\bibitem{yan20222dpassb}
Yan, X., Gao, J., Zheng, C., Zheng, C., Zhang, R., Cui, S., Li, Z.: {{2DPASS}}:
  {{2D Priors Assisted Semantic Segmentation}} on {{LiDAR Point Clouds}}. In:
  Eur. Conf. Comput. Vis. (2022)

\bibitem{ye2022lidarmultinet}
Ye, D., Chen, W., Zhou, Z., Xie, Y., Wang, Y., Wang, P., Foroosh, H.:
  {{LidarMultiNet}}: {{Unifying LiDAR Semantic Segmentation}}, {{3D Object
  Detection}}, and {{Panoptic Segmentation}} in a {{Single Multi-task
  Network}}. In: IEEE Conf. Comput. Vis. Pattern Recog. Worksh. (2022)

\bibitem{ye2023lidarmultinet}
Ye, D., Zhou, Z., Chen, W., Xie, Y., Wang, Y., Wang, P., Foroosh, H.:
  {LidarMultiNet: {{Towards}} a Unified Multi-Task Network for Lidar
  Perception}. In: Conf. AAAI Artif. Intell. vol.~37, pp. 3231--3240 (2023)

\bibitem{ye2022efficient}
Ye, M., Wan, R., Xu, S., Cao, T., Chen, Q.: Efficient {{Point Cloud
  Segmentation}} with {{Geometry-Aware Sparse Networks}}. In: Avidan, S.,
  Brostow, G., Ciss{\'e}, M., Farinella, G.M., Hassner, T. (eds.) Eur. Conf.
  Comput. Vis. pp. 196--212. {Springer} (2022)

\bibitem{zhang2022visualizing}
Zhang, C., Duc, K.D., Condon, A.: {Visualizing {{DNA}} Reaction Trajectories
  with Deep Graph Embedding Approaches}. In: Adv. Neural Inform. Process. Syst.
  Worksh. {arXiv} (2022)

\bibitem{zhao2021point}
Zhao, H., Jiang, L., Jia, J., Torr, P., Koltun, V.: {Point Transformer}. In:
  Int. Conf. Comput. Vis. (2021)

\bibitem{zhu2021cylindrical}
Zhu, X., Zhou, H., Wang, T., Hong, F., Ma, Y., Li, W., Li, H., Lin, D.:
  Cylindrical and {{Asymmetrical 3D Convolution Networks}} for {{LiDAR
  Segmentation}}. In: IEEE Conf. Comput. Vis. Pattern Recog. (2021)

\bibitem{zhuang2021perception}
Zhuang, Z., Li, R., Jia, K., Wang, Q., Li, Y., Tan, M.: {Perception-Aware
  Multi-Sensor Fusion for {{3D LiDAR}} Semantic Segmentation}. In: Int. Conf.
  Comput. Vis. pp. 16280--16290 (2021)

\end{thebibliography}
\end{document}